\documentclass[letterpaper]{article} 
\usepackage{aaai25}
\usepackage{times}  
\usepackage{helvet}  
\usepackage{courier}  
\usepackage[hyphens]{url}  
\usepackage{graphicx} 
\usepackage{natbib}  
\usepackage{caption} 
\usepackage{graphicx}
\usepackage{subcaption}
\usepackage{multirow}
\usepackage{amsmath} 
\usepackage{algorithm}
\usepackage{algorithmic}
\usepackage{xcolor}
\usepackage{rotating}
\usepackage{graphicx}
\usepackage{fancyvrb}
\usepackage{newfloat}
\usepackage{listings}

\DeclareCaptionStyle{ruled}{labelfont=normalfont,labelsep=colon,strut=off} 
\floatstyle{ruled}
\newfloat{listing}{tb}{lst}{}
\floatname{listing}{Listing}
\title{LNS2+RL: Combining Multi-Agent Reinforcement Learning with Large Neighborhood Search in Multi-Agent Path Finding}
\author {
    Yutong Wang\textsuperscript{\rm 1},
    Tanishq Duhan\textsuperscript{\rm 1},
    Jiaoyang Li\textsuperscript{\rm 2},
    Guillaume Sartoretti\textsuperscript{\rm 1}
}
\affiliations {
    \textsuperscript{\rm 1}Department of Mechanical Engineering, National University of Singapore, Singapore\\
    \textsuperscript{\rm 2}Robotics Institute, Carnegie Mellon University, USA\\
    e0576114@u.nus.edu, e1280621@u.nus.edu, jiaoyangli@cmu.edu, guillaume.sartoretti@nus.edu.sg
}

\begin{document}

\maketitle

\begin{abstract}
Multi-Agent Path Finding (MAPF) is a critical component of logistics and warehouse management, which focuses on planning collision-free paths for a team of robots in a known environment.
Recent work introduced a novel MAPF approach, LNS2, which proposed to repair a quickly obtained set of infeasible paths via iterative replanning, by relying on a fast, yet lower-quality, prioritized planning (PP) algorithm.
At the same time, there has been a recent push for Multi-Agent Reinforcement Learning (MARL) based MAPF algorithms, which exhibit improved cooperation over such PP algorithms, although inevitably remaining slower.
In this paper, we introduce a new MAPF algorithm, LNS2+RL, which combines the distinct yet complementary characteristics of LNS2 and MARL to effectively balance their individual limitations and get the best from both worlds.
During early iterations, LNS2+RL relies on MARL for low-level replanning, which we show eliminates collisions much more than a PP algorithm.
There, our MARL-based planner allows agents to reason about past and future information to gradually learn cooperative decision-making through a finely designed curriculum learning.
At later stages of planning, LNS2+RL adaptively switches to PP algorithm to quickly resolve the remaining collisions, naturally trading off solution quality (number of collisions in the solution) and computational efficiency.
Our comprehensive experiments on high-agent-density tasks across various team sizes, world sizes, and map structures consistently demonstrate the superior performance of LNS2+RL compared to many MAPF algorithms, including LNS2, LaCAM, EECBS, and SCRIMP.
In maps with complex structures, the advantages of LNS2+RL are particularly pronounced, with LNS2+RL achieving a success rate of over 50\% in nearly half of the tested tasks, while that of LaCAM, EECBS and SCRIMP falls to 0\%.
\end{abstract}
\begin{links}
    \link{Code}{https://github.com/marmotlab/LNS2-RL}
\end{links}

\section{Introduction}
Multi-Agent Path Finding (MAPF) focuses on the problem of generating a set of shortest collision-free paths for a team of agents from their initial positions to designated goals in a shared environment.
It is a core problem in many applications, such as warehouse automation, where balancing solution quality and speed is crucial.
A recent paper~\cite{li2022mapf} introduced LNS2, a large neighborhood search-based MAPF algorithm that iteratively replans currently infeasible paths, differing from mainstream MAPF algorithms that rely on more expensive search/backtracking.
LNS2 exhibits greater scalability than systematic search algorithms~\cite{sharon2015conflict} and prioritized planning algorithms~\cite{erdmann1987multiple}.
It first quickly generates an initial solution that may contain collisions, and then iteratively replans for carefully selected subsets of agents, using a fast, low-level Prioritized Planning algorithm based on Safe Interval Path Planning with Soft Constraints (termed PP+SIPPS) to finally yield a collision-free solution.
However, in challenging tasks, the paths generated by PP+SIPPS still often contain a large number of collisions, inevitably increasing the cumulative time needed to address the overall MAPF problem.

In parallel to these developments, recent years have seen the development of many (slower) MAPF planners based on Multi-Agent Reinforcement Learning (MARL), which exhibit enhanced cooperation compared to prioritized planning algorithms.
However, MARL-based planners inevitably exhibit performance degradation when deviating from their training setup (usually only 8 agents).
Moreover, the one-shot Success Rate (SR) of MARL-based planners often drops sharply due to a few agents failing to reach their goals within the timestep limit, thereby wasting high-quality paths for the majority that succeeded.

In this paper, we propose combining large neighborhood search and MARL, introducing a new algorithm called LNS2+RL, which balances the drawbacks of both LNS2 and MARL-based planners.
The overall framework of LNS2+RL follows that of LNS2, iteratively updating the overall solution with re-planned paths.
However, during the challenging early stages of iterative replanning, it replaces PP+SIPPS with a MARL-based planner, generating paths with fewer collisions at a slower speed.
There, we rely on SIPPS to fix the paths planned by agents unable to reach their goals within a limited number of timesteps, thus ensuring the completeness of our MARL-based planner and maximizing the utilization of existing high-quality paths.
LNS2+RL then adaptively switches to the original PP+SIPPS to swiftly clean up remaining collisions, naturally trading-off solution quality (number of collisions in the solution) and replanning speed.
Moreover, since the number of agents involved in each replanning task (i.e., neighborhood size) is fixed to 8, LNS2+RL can naturally scale the overall planning to thousands of agents while keeping encountered tasks close to MARL's training distribution.

We trained only a single MARL model but tested LNS2+RL on an extensive set of high-agent-density tasks, where the team sizes and map structures differed from those used during training.
Our results indicate that LNS2+RL consistently outperforms LNS2 in SR across nearly all tasks, maintaining a comparable Sum of Costs (SoC), and notably reducing the number of remaining Colliding Pairs (CP) in the overall solution.
Additionally, LNS2+RL surpasses the search-based bounded-suboptimal algorithm EECBS~\cite{li2021eecbs} and MARL-based planner SCRIMP~\cite{wang2023scrimp} in SR across all tasks.
Compared to the search-based unbounded-suboptimal algorithm LaCAM~\cite{okumura2023lacam}, LNS2+RL exhibits a substantial advantage in SR on maps with complex obstacle structure (achieving 100\% SR compared to LaCAM's 0\%), while also achieving significantly lower SoC on maps with simple obstacle structures. 
Finally, we experimentally validated our planner in a hybrid simulation of a warehouse environment under real-world conditions.

\section{Prior Work}

MAPF algorithms developed in recent decades fall into three main categories based on solution optimality: optimal, bounded-suboptimal, and unbounded-suboptimal. 
Common optimal algorithms include Conflict-Based Search (CBS)~\cite{sharon2015conflict}, joint-state A*, and their variants with additional speed-up techniques such as ICBS~\cite{boyarski2015icbs}, M*~\cite{wagner2011m}, EPEA*~\cite{goldenberg2014enhanced}, among others.
Bounded-suboptimal algorithms trade off solution quality and runtime by finding a solution whose cost is at most $w$ times the optimal cost, where $w$ is a user-specified suboptimality factor.
Such MAPF algorithms are usually variants of optimal MAPF algorithms, including EECBS~\cite{li2021eecbs}, and inflated M*~\cite{wagner2015subdimensional}, etc.
Both optimal and bounded-suboptimal algorithms are able to ensure high solution quality but face scalability challenges when handling thousands of agents, often encountering timeouts or memory overflows.
Conversely, unbounded-suboptimal algorithms (such as prioritized planning~\cite{erdmann1987multiple} and rule-based algorithm~\cite{luna2011push}) prioritize speed over optimality, scaling to thousands of agents easily, but may yield substantially suboptimal solutions in complex tasks and often do not guarantee completeness. 

More recently, learning-involved approaches have begun to emerge. 
One class of approaches, exemplified by PRIMAL~\cite{sartoretti2019primal}, leverages MARL or imitation learning to directly train MAPF planners. 
These approaches often incorporate non-learning algorithms to further improve the performance of the learning-based planners by enriching observations~\cite{damani2021primal}, designing rewards~\cite{skrynnik2024learn}, or generating training data~\cite{li2021message}.
Although these learning-based planners emphasize their superior scalability compared to optimal or bounded-suboptimal search-based algorithms, nearly all of them still fail to outperform scalable suboptimal non-learning algorithms like LNS2 within the same time constraints.
Another class of methods seeks to enhance conventional algorithms by using machine learning techniques to replace manually crafted heuristics or components, exemplified by learning a priority ordering for prioritized planning~\cite{zhang2022learning}, selecting subsets of agents that need to be replanned~\cite{huang2022anytime}.
However, these approaches do not directly optimize the path planner itself.

\section{Background}
\subsection{Multi-Agent Path Finding}
In this paper, we consider a common formulation of the MAPF problem, defined within a 2D 4-neighbor grid with a finite set of static obstacles, and encompassing a set of $m$ agents $A = \{a_1, \ldots, a_m\}$. 
Each agent $a_i$ initiates from a distinct start $s_i$, with the objective of reaching its designated goal $g_i$. 
At each discretized timestep, an agent can either move to an adjacent cell not occupied by an obstacle or wait at its current cell.
The \emph{path} $p_i$ with length $k$ (where the path length $k$ may vary for different paths) for each agent is a sequence of adjacent or identical cells $(c_1^{i}, \ldots, c_k^{i})$ satisfying $c_1^{i} = s_i$, $c_k^{i} = g_i$.
We assume that the agents wait at their goals until all agents have reached their goals.
A \emph{solution} of MAPF $P=\left\{p_i \mid a_i \in A\right\}$ is a path set for all agents, ensuring all agents reach their goals without any vertex (agents occupy the same cell at the same timestep) or swapping collisions (agents simultaneously attempting to move into each other’s current cells).
This paper adopts the \emph{sum of costs} metric as one of the evaluation criteria, which is defined as the sum of timesteps required by each agent to reach its goal $\sum_{1 \leq i \leq m}\left|p_i\right|$.

\subsection{LNS2}
LNS2~\cite{li2022mapf} is an unbounded-suboptimal and memory-efficient MAPF algorithm based on large neighborhood search.
It begins by planning a set of paths that may contain collisions, and then iteratively selects subsets for replanning using adaptive neighborhood selection based on past experience.
A replanning solver then computes new paths for these selected agents and evaluates whether to adopt the new paths depending on whether they reduce or maintain the number of CP within the overall solution.
This replanning process continues iteratively until the overall path set is collision-free or the time limit is exceeded. 

The low-level replanning solver employed in LNS2 is a prioritized planning algorithm PP+SIPPS, which utilizes SIPPS to sequentially plan paths for each agent based on a randomly assigned priority order, where lower-priority agents need to avoid the paths already planned for higher-priority agents as much as possible.
SIPPS is designed to find a path for an agent that avoids collisions with hard obstacles (static obstacles) while minimizing collisions with soft obstacles (pre-existing paths).
It ensures completeness and returns the shortest path with zero soft collisions if one exists. 
However, in dense-obstacle scenarios, SIPPS may produce a path with significantly more soft collisions than the minimum due to its heuristic function ignoring soft collisions that occur when the agent waits within a safe interval (a contiguous period of time) that contains soft obstacles.
More importantly, the prioritized planning algorithm relies on a static-priority-based conflict avoidance mechanism, which significantly reduces the flexibility of the planning process.
This limitation results in suboptimal paths and a larger number of collisions for lower-priority agents, making it difficult to achieve global cooperation.
Despite the fast speed of PP+SIPPS, this reduced solution quality (more collisions) necessitates a higher number of iterations to fully solve high-difficulty MAPF instances, resulting in substantial cumulative time expenditure.
Therefore, there is an urgent need to develop a new replanning solver that can generate high-quality solutions (fewer collisions) while potentially sacrificing speed, to find a better balance between quality and time throughout the overall planning.

\subsection{Path Finding with Mixed Dynamic Obstacles}
The replanning task of LNS2 for selected agents can be defined as the Path Finding with Mixed Dynamic Obstacles (PMDO) problem. 
In the PMDO problem, two finite sets of obstacles are considered. 
The first category, hard obstacles, aligns with the immovable obstacles in conventional MAPF. 
The second category, soft obstacles, refers to other unselected agents and is characterized by the ability to move to any cell not occupied by hard obstacles and remain there for an arbitrary duration.
Notably, the complete path of each soft obstacle is fully disclosed and remains constant throughout the replanning task.
Any occupied cell at a given timestep $t$ can contain either a single hard obstacle or multiple soft obstacles/agents.
The objective of the replanning task is to find a path for each selected agent from its start to its goal that avoids collisions with hard obstacles, prioritizes minimizing collisions with soft obstacles and other selected agents, and then minimizes the path length.

\begin{algorithm}[tb]
\caption{LNS2+RL}
\label{alg:LNS2+RL}
\begin{algorithmic}[1]
   \STATE {\bfseries Input:} A MAPF instance $I$
    \STATE $P =\left\{p_i \mid a_i \in A\right\} \leftarrow$ lnitialSolver $(I)$
   \STATE $Q = \emptyset$
   \STATE $t_m = \max\{|p_i| : p_i \in P\} $, $t_l = d_l \cdot t_m$, $t_h = d_h \cdot t_m$
   \WHILE{runtime limit not exceeded and $CP(P)> 0$}
       \STATE $A_s, P^{-} \leftarrow \operatorname{selectAgentSet}(I, P)$
        \IF {$|Q| < \mu$ or $\bar{Q} > \rho$}
            \STATE $P^{rl}, \hat{P} \leftarrow \operatorname{MARLPlanner}\left(I, P \setminus P^{-}, A_s, t_l \right)$
            \STATE $A_r \leftarrow \{a_i \in A_s : p_{i, \text{end}}^{rl}  \neq g_i\}$
            \STATE $P^{rl} \leftarrow \operatorname{PrioritizedPlanner}\left(I, P \setminus P^{-}, P^{rl}, A_r, t_l\right)$
            \STATE $A_h \leftarrow \{a_i \in A_s : |p_i^{rl}| \geq t_h\}$
            \STATE $P^{rl} \leftarrow \operatorname{PrioritizedPlanner}\left(I, P \setminus P^{-}, P^{rl}, A_h, 0\right)$
            \STATE Update $Q$ according to if $CP(P^{rl})$ $\geq$ $CP(\hat{P})$
            \IF{$ CP(\left(P \backslash P^{-}\right) \cup P^{rl}) \le CP(P)$}
                \STATE $P \leftarrow\left(P \backslash P^{-}\right) \cup P^{rl}$
                 \STATE \textbf{continue}
            \ENDIF
        \ELSE
             \STATE $\hat{P} \leftarrow \operatorname{PrioritizedPlanner}\left(I, P \setminus P^{-}, \emptyset, A_s, 0\right)$
        \ENDIF    
    \IF{$ CP(\left(P \backslash P^{-}\right) \cup \hat{P}) \le CP(P)$}
        \STATE $P \leftarrow\left(P \backslash P^{-}\right) \cup \hat{P}$
        \ENDIF  
   \ENDWHILE
\STATE \textbf{return} $P$
\end{algorithmic}
\end{algorithm}

\section{Method: LNS2+RL}
In this section, we first present the overall framework of LNS2+RL and then detail the MARL model training.

\subsection{Overall Framework}
The strengths and weaknesses of PP+SIPPS and our MARL-based planner are distinct yet complementary. 
PP+SIPPS is fast and effective for simple tasks but tends to generate paths with more soft collisions in complex tasks.
Conversely, our MARL-based planner, although slower (approximately seven times slower), excels at planning paths with fewer collisions in difficult tasks (about one-fourth of PP+SIPPS, with detailed data provided in ``Performance analysis`` subsection). 
Therefore, LNS2+RL algorithm retains the overarching framework of LNS2 but uses the MARL-based planner in the initial, more challenging stages of iterative planning for better solution quality (fewer collisions). 
It then adaptively switches to PP+SIPPS to quickly clean up the remaining collisions, thus balancing solution quality (number of collisions in the solution) with the speed of the replanning task solvers. 
Moreover, the one-shot SR of MARL-based planners often drops sharply because a few agents fail to reach their goals within the constrained episode length, even though most agents have already found high-quality paths.
To address this, we employ PP+SIPPS to supplement paths for these few agents, ensuring the completeness of the final solution.

The pseudocode for the LNS2+RL algorithm is presented in Algorithm~\ref{alg:LNS2+RL}.
Specifically, for solving a MAPF instance $I$, LNS2+RL starts by invoking the initial solver (PP+SIPPS) to generate a path set $P$ containing collisions (lines 2).
Furthermore, LNS2+RL introduces a queue $Q$ with a maximum capacity of $\mu$, used to store the comparison results of the number of CP in the paths generated by PP+SIPPS and the MARL planner, enabling adaptive switching (line 3).
Subsequently, LNS2+RL counts the maximum length of individual paths in the initial solution ($t_m$), sets the timestep threshold $t_l$ for pausing the MARL-planner, along with the maximum allowable path length $t_h$, where the constant value $d_h > d_l$ (line 4). 
If the termination criteria are not met, LNS2+RL iteratively selects a subset of agents $A_s \subseteq A$ using adaptive neighborhood selection (the same as in LNS2) and removes their paths $P^{-} = \left\{p_i \in P: a_i \in A_s\right\}$ from $P$, treating the remaining part of the path set $P \backslash P^{-}$ as fixed to construct a PMDO replanning task (lines 6).
The neighborhood size (the size of  $A_s$) is fixed at 8 (equivalent to the number of agents used in MARL training).
If the size of $Q$ is below $\mu$, or its average $\bar{Q}$ exceeds another threshold $\rho$, LNS2+RL infers that the benefit of using the MARL-planner exceeds that of PP+SIPPS at this stage (line 7).
Therefore, LNS2+RL chooses the trained MARL planner as the replanning solver. 
The function in line 8 uses the MARL-planner to generate actions until timestep $t_l$ (or until the replanning task is fully resolved, i.e., all agents have reached their goal without future soft collisions).
It yields two path sets: $\hat{P}$, derived by PP+SIPPS at the onset of the replanning task, acting as the reference paths for the MARL policy, and path set $P^{rl}$, computed by the MARL policy.
For incomplete paths in $P^{rl}$ where agents haven't reached their goals (Line 9), we employ PP+SIPPS to plan the remaining paths from $t_l$ and merge these new paths with the existing ones (Line 10).
Subsequently, all path lengths in $P^{rl}$ are reviewed, and those exceeding $t_h$ are directly replaced with paths re-planned by PP+SIPPS from timestep zero (lines 11-12).
This operation aims to remove non-contributory segments planned by the MARL planner, thus reducing the SoC of the final solution $P$ and planning time for subsequent iterations.
Next, LNS2+RL updates $Q$ by appending 0 if $CP(P^{rl}) \geq CP(\hat{P})$, and 1 otherwise (lines 13).
Finally, LNS2+RL first checks whether $P^{rl}$ meets the update condition (line 14), namely if the CP of the repaired path set $\left(P \backslash P^{-}\right) \cup P^{rl}$ is no larger than that of the old path set.
If not, LNS2+RL continues to check whether $\hat{P}$ meets the update condition (line 21).
The MARL planner continues to re-plan until either the switching condition or termination condition is met.
Once the replanning solver switches to PP+SIPPS, it cannot switch back to the MARL-planner and always addresses remaining collisions using the same procedure as LNS2.

\subsection{Replanning Tasks as a MARL Problem}
\label{sec:marl}
\subsubsection{RL Environment Setup}
Aligned with previously outlined MAPF and PMDO problems, we consider a discrete 2D 4-neighbor grid environment where the size of each agent, goal, and obstacle is one grid cell. 
The agents' actions that result in collisions with hard obstacles or exceed world boundaries are classified as invalid.
Only valid actions are sampled during training and evaluation, and an additional supervised loss is added to the overall MARL loss to reduce the probability of choosing invalid actions.

We initially establish a MAPF task comprising $m$ agents, and use PP+SIPPS to compute the initial path set $P$ for these agents.
Subsequently, a subset of agents is selected, and during the training phase of the MARL model, we keep the subset size $n$ (i.e., the neighborhood size) to 8.
The PMDO replanning task is then formulated by designating the chosen subset as the controlled agents of the MARL model while the remaining agents act as soft obstacles.
Throughout the remainder of this section, we refer to the selected agents simply as agents and to the others as soft obstacles. 
A training episode terminates either when all agents are on goal at the end of a timestep (and the timestep is larger than the maximum length of the soft obstacle paths) or when the episode reaches the pre-defined timestep limit.

\begin{table*}[]
\centering
\scalebox{0.8}{
\begin{tabular}{c|c|c|c|c|c|c|c}
\hline
Action & \begin{tabular}[c]{@{}c@{}}Move/Stay  (not on goal)\end{tabular} & \begin{tabular}[c]{@{}c@{}}Stay  (on goal)\end{tabular} & \begin{tabular}[c]{@{}c@{}}Exceed max length of $\hat{P}$\end{tabular} & Revisit  & \begin{tabular}[c]{@{}c@{}}Collision with obstacles\end{tabular} & \begin{tabular}[c]{@{}c@{}}Avoid congestion\end{tabular} & Off-route \\ \hline
Reward & {[}-0.4,-0.5,-0.6{]} & 0.0 & -0.2 & -0.4 & -1.5 & Eq. \ref{eq:uti_reward}& Eq.\ref{eq:off_reward} \\ \hline
\end{tabular}}
\caption{Reward structure.}
\label{table:reward}
\end{table*}

\subsubsection{Observation and Reward}
Aligned with previous works~\cite{sartoretti2019primal}, the environment is assumed to be partially observable for each agent, restricting it to only access information within its self-centered field of view (FoV).
This assumption enables the trained policy to scale to arbitrary world sizes while fixing the input dimensions of the neural network.

Each agent's observation is divided into two parts.
Here, we take agent $i$ and timestep $t$ as an example.
The first part of the observation is an eight-element vector consisting of the normalized distance between $c_t^i$ (the position of agent $i$ at timestep $t$) and $g_i$ along the x-axis and y-axis, as well as the total Euclidean distance, the ratio of the number of CP in the path planned so far by the MARL-planner compared to that in $\hat{p_i}$ (agent i’s SIPPS path), the ratio of $t$ to the predefined episode length, the ratio of $t$ to the maximum length of $\hat{P}$, the ratio of agents who have reached goal, and agent $i$'s action at $t-1$.
The section part of the observation consists of thirty-one 2D matrices (Summary of these metrics is in the supplementary material).
These matrices can be classified into four categories.
\paragraph{1. Design from prior work}
Building on previous research~\cite{wang2023scrimp}, we introduce four matrices to represent the positions of hard obstacles, other agents, observable agents' goals, and agent $i$'s own goal. 
Additionally, there are four binary matrices (one for each move action) that indicate whether selecting the action will bring agent $i$ closer to its goal, as determined by the backward Dijkstra algorithm.

\paragraph{2. Reference path}
To better guide agent $i$ closer to its goal, we introduce the \textit{SIPPS map}, which represents all cells traversed by agent $i$'s SIPPS path $\hat{p}_i$ (computed only at the beginning of the replanning task) during timesteps $t-15$ to $t+15$.
To motivate agents to refer to the SIPPS paths while allowing necessary detours to avoid collisions, we define an off-route penalty $ro_t^i$: 
\begin{equation}
\label{eq:off_reward}
   ro_t^{i}=-\alpha \min \left\|c_t^i-c\right\|_2 , {c \in \left\{\left(\hat{p}_i\right)_j \mid t-15 \leq j < t+15\right\}},
\end{equation}
where $\alpha$ is a predefined hyperparameter, $c$ represents all cells contained in the SIPPS map of agent $i$.

\paragraph{3. Avoid collisions}
To avoid collisions with hard and soft obstacles (with known paths), we further introduce \textit{future path maps of soft obstacles}, and an \textit{occupancy duration map} along with its complementary \textit{blank duration map}.
The future path maps, comprising nine matrices, represent the positions of soft obstacles from $t$ to $t+8$. 
The occupancy duration map denotes the ratio of time a cell will be occupied by obstacles to the predefined episode termination length, while the blank duration map reflects the ratio of time a cell will remain obstacle-free.

The future paths of other agents are crucial to enhancing cooperation, but these paths are unknown in advance.
Hence, we introduce \textit{predicted path maps of other agents} and employ a straightforward yet efficient method for forecasting them.
To predict the future path of agent $j$, we first calculate the spatial Euclidean distance between $c_t^{j}$ and all cells on $\hat{p}_j$, and treat the absolute difference of their timestep as the temporal distance.
Next, we compute a weighted sum of the spatial and temporal distances for each cell, with weights of 0.9 for spatial distance and 0.1 for temporal distance.
If the cell with the smallest weighted sum can be reached from $c_t^{j}$ in a single step, it is considered the predicted position at $t+1$; otherwise, agent $j$ is assumed to remain stationary, and its current position $c_t^{j}$ is regarded as the predicted position at $t+1$.
Subsequently, the next four cells on $\hat{p}_j$ starting from the cell with the smallest weighted sum are designated as the predicted positions from $t+2$ to $t+5$.

\paragraph{4. Alleviate congestion}
In path planning tasks involving a large number of agents, congestion often arises in bottleneck areas of the map, where all agents attempt to traverse the same set of cells, leading to under-utilization of other cells and significantly increased collision risks.
To alleviate congestion, we introduce the \textit{cell utilization map} and the \textit{direction utilization maps}, which visualize congestion and assist agents in identifying and avoiding high-density areas, promoting more evenly dispersed paths.
Specifically, denote $U (c)$ and $U (c_1, c_2)$ as the usage of cell $c$ and direction $(c_1, c_2)$ during the time period from $t-2$ to $t+15$.
They represent, respectively, the total timesteps during that time period in which soft obstacles and agents occupy cell $c$ and traverse direction $(c_1, c_2)$.
The normalized value $10 \frac{U (c)}{m}$ and $10 \frac{U (c_1, c_2)}{m}$ are then used as the value of each grid in the cell utilization map and four direction utilization maps.
Moreover, we further encourage agents to avoid these congested areas by giving them a small penalty $rc_t^{i}$ calculated by  
\begin{equation}
\label{eq:uti_reward}
   rc_t^{i} = 10 (\delta_c \frac{U (c)}{m} +\delta_d \frac{U (c_1, c_2)}{m}) .
\end{equation}
Here $\delta_c = 0.225$ and $\delta_d = 0.075$ are used to balance between cell and direction information.

Our reward structure is shown in Table \ref{table:reward}.
To expedite task completion, agents currently off-goal are penalized at each step, with its magnitude scaling up according to the task's difficulty within the curriculum learning framework (refer to ``Training process`` subsection for details).
Agents receive an extra penalty if the current timestep exceeds the maximum length of the current $\hat{P}$.
Additionally, to foster exploration and learn effective policies, agents are penalized for revisiting their previous timestep's location.
Furthermore, we employ reward shaping based on the distance from the goal to guide learning while preserving the invariance of the optimal policy.

\subsubsection{Network Structure}
\begin{figure}[h]
  \centering
  \includegraphics[width=1\linewidth]{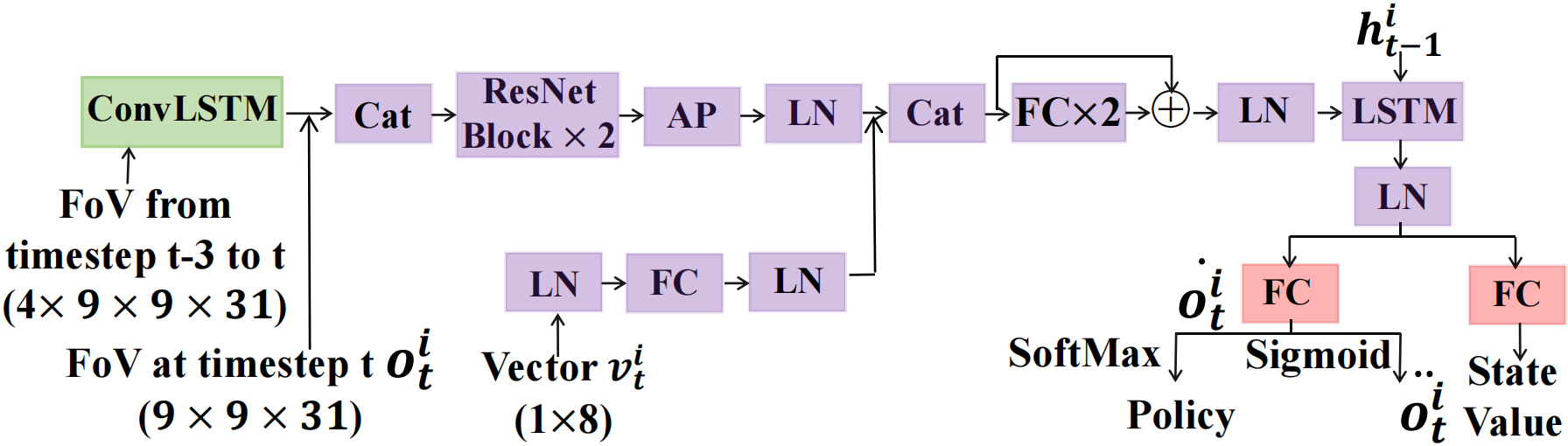}
  \caption{Network structure.
AP, Cat, LN, and FC represent average pooling, concatenation, layer normalization, and fully connected layer, respectively.
$h_{t-1}^i$ represents the hidden state output by the LSTM unit at the previous time step.}
  \label{fig:net}
\end{figure}

Figure~\ref{fig:net} illustrates the network architecture of our MARL model.
Given our assumption of a homogeneous team, we rely on weight sharing across all agents, enabling the model to scale effectively with the number of agents during evaluation.

Understanding the underlying dynamics of the environment is crucial in PMDO tasks.
Therefore, we additionally incorporate observations from the past three time steps as input to provide a more comprehensive understanding of how the environment changes over time. 
The aggregated observation first passes through a Convolutional Long Short-Term Memory (ConvLSTM)~\cite{shi2015convolutional} unit capable of capturing spatial and temporal relationships and then concatenates its last hidden state with the current observation $o_t^i$.
The concatenated data and the eight-element vector $v_t^i$ are processed through multiple neural network layers to generate a vector $\dot{o}_t^i$ for imitation learning and a predicted state value.
Subsequently, $\dot{o}_t^i$ passes through a softmax activation to derive the agent's policy and a sigmoid activation to yield a vector $\ddot{o}_t^i$ for supervised invalid action loss.

\begin{figure*}[t]
\centering
\begin{subfigure}[b]{0.2\textwidth}
    \includegraphics[width=\textwidth]{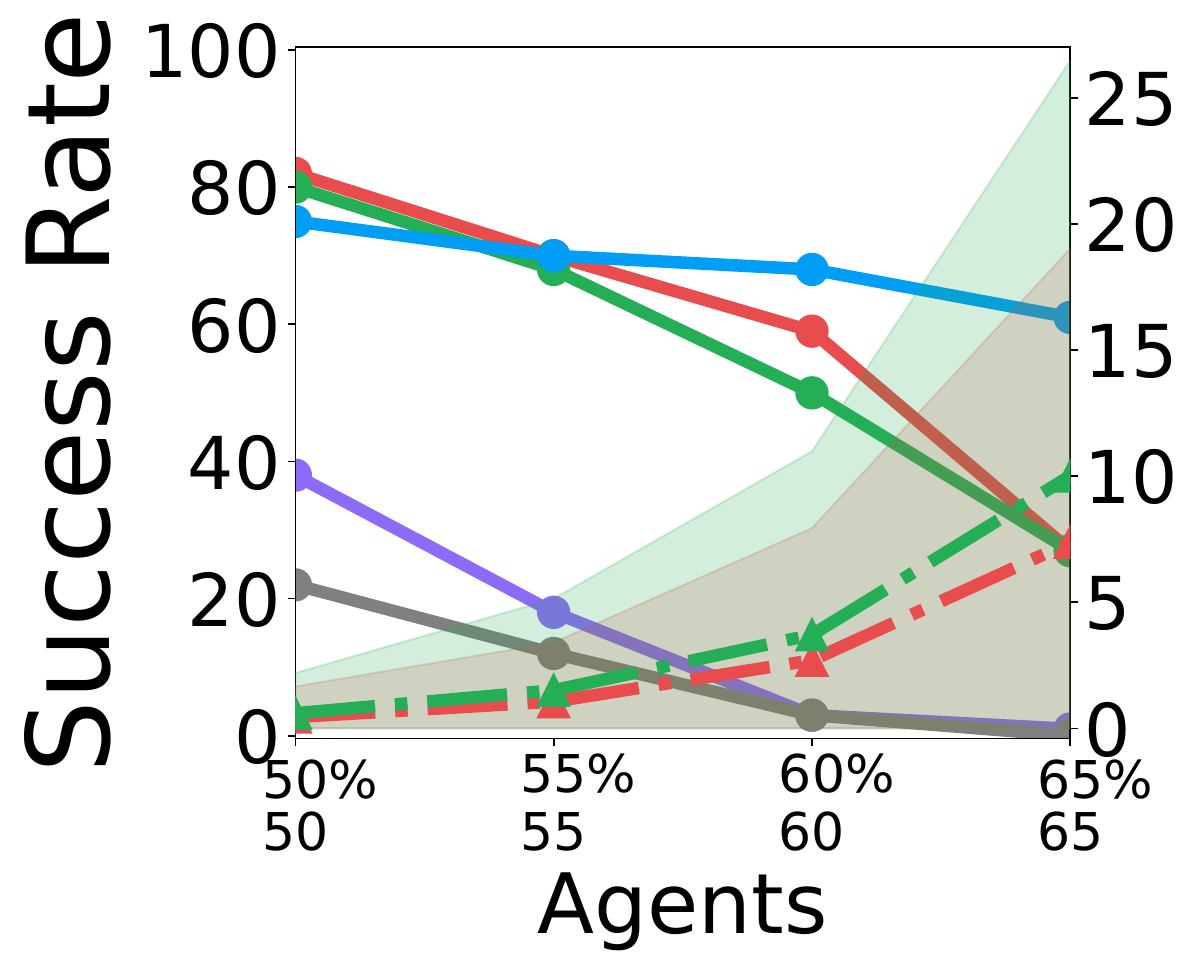}
    \caption{Random-small}
    \label{fig:Random-small}
\end{subfigure}
\begin{subfigure}[b]{0.2\textwidth}
    \includegraphics[width=\textwidth]{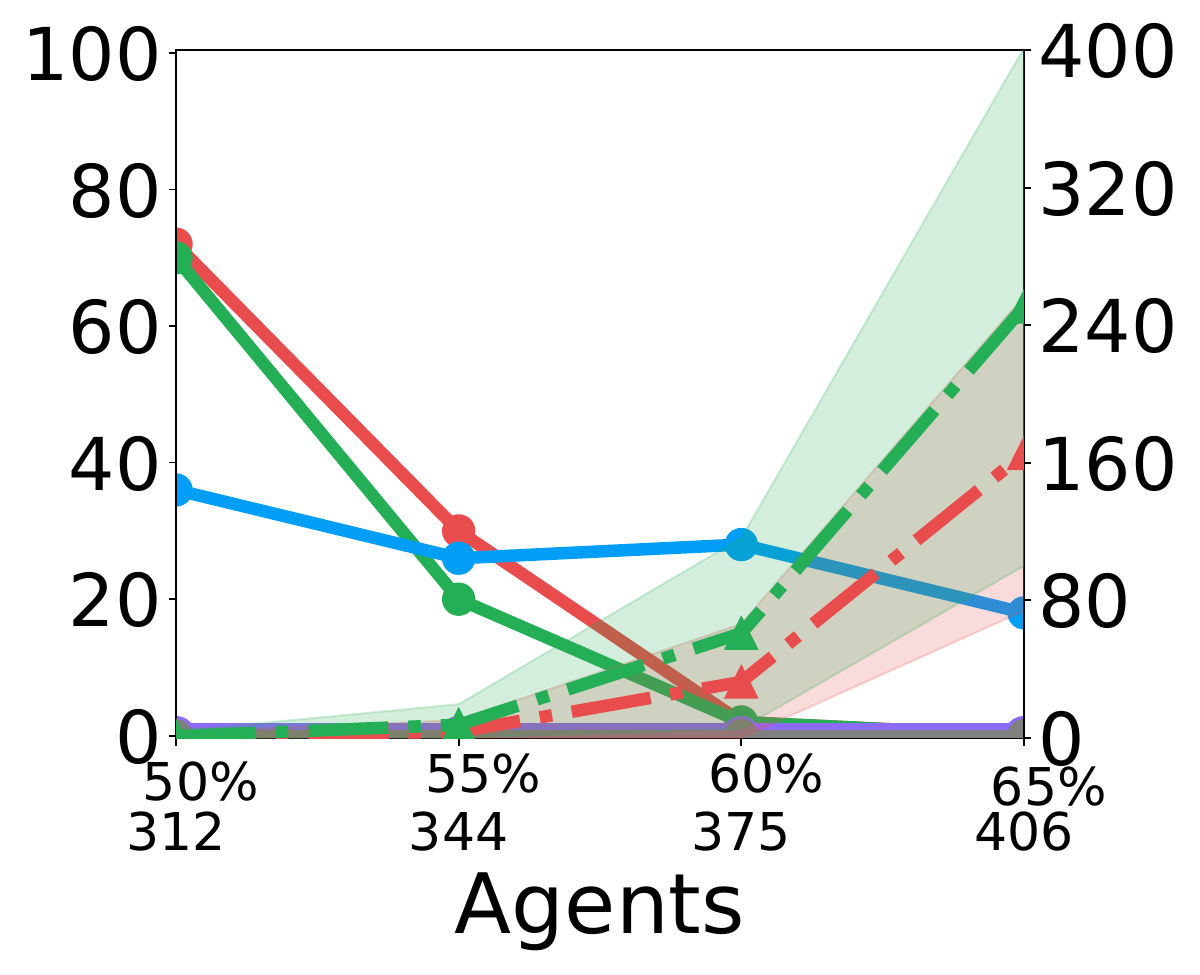}
    \caption{Random-medium}
    \label{fig:Random-medium}
\end{subfigure}
\begin{subfigure}[b]{0.2\textwidth}
    \includegraphics[width=\textwidth]{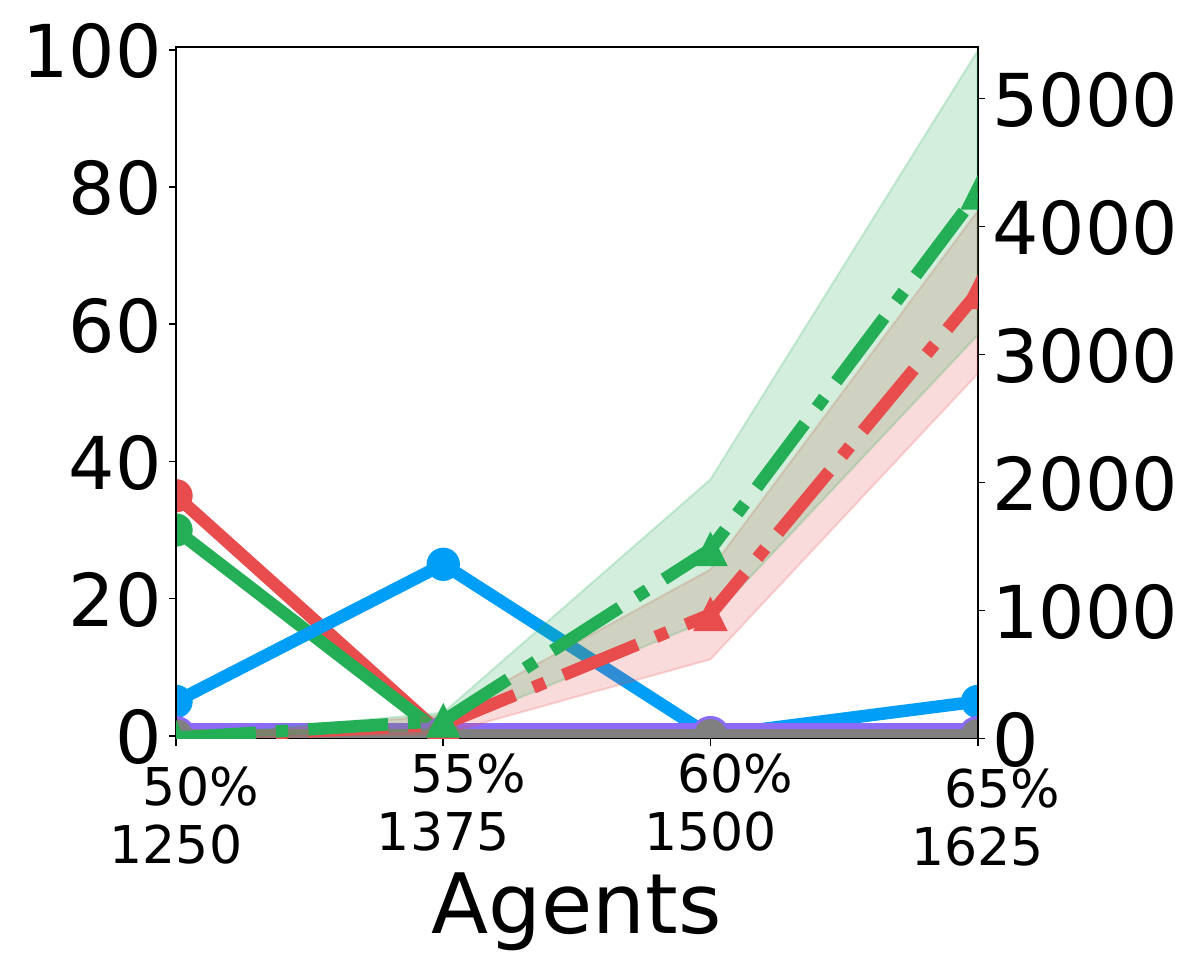}
    \caption{Random-large}
    \label{fig:Random-large}
\end{subfigure}
\begin{subfigure}[b]{0.2\textwidth}
    \includegraphics[width=\textwidth]{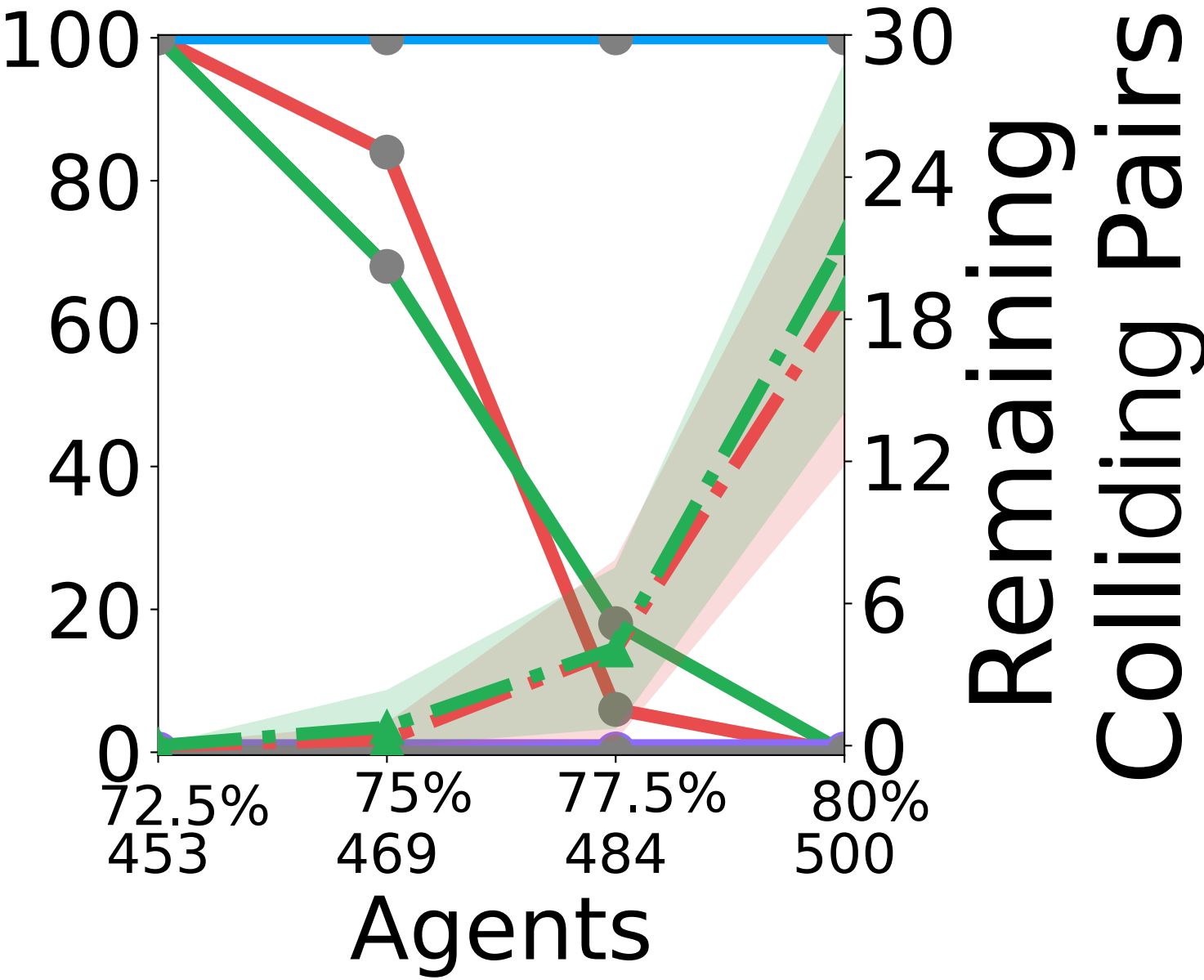}
    \caption{Empty}
    \label{fig:Empty}
\end{subfigure}
\begin{subfigure}[b]{0.2\textwidth}
    \includegraphics[width=\textwidth]{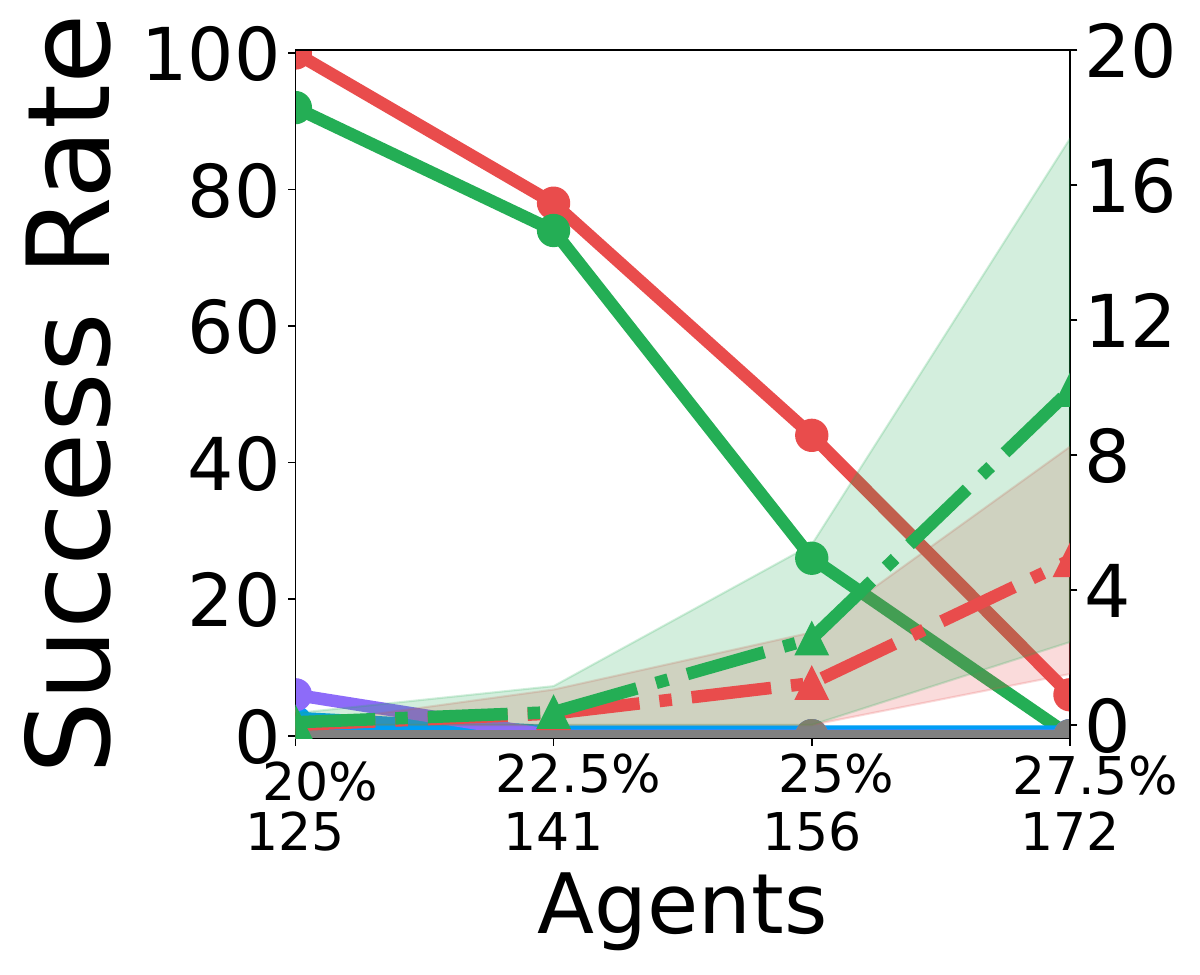}
    \caption{Maze}
    \label{fig:Maze}
\end{subfigure}
\begin{subfigure}[b]{0.2\textwidth}
    \includegraphics[width=\textwidth]{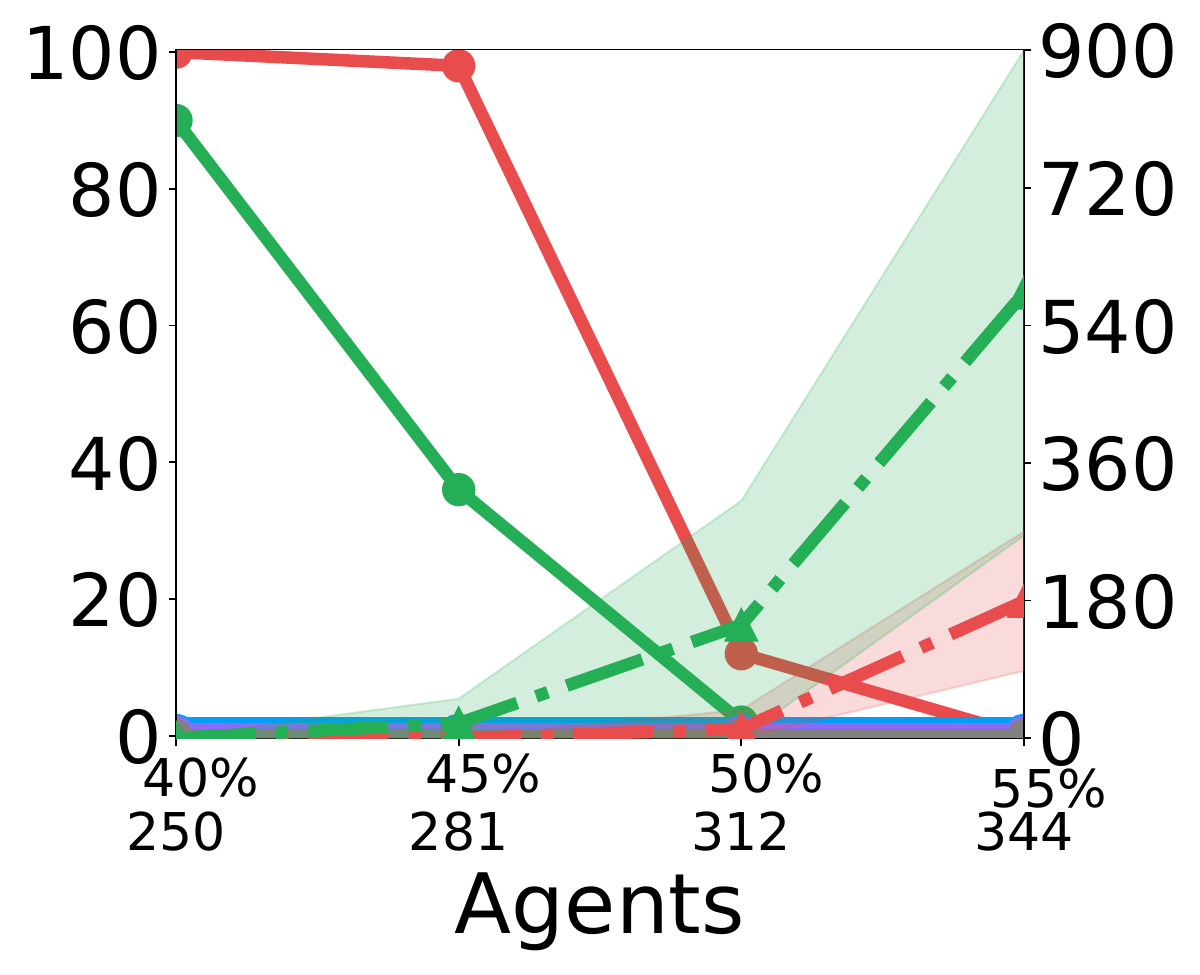}
    \caption{Room}
    \label{fig:Room}
\end{subfigure} 
\begin{subfigure}[b]{0.2\textwidth}
    \includegraphics[width=\textwidth]{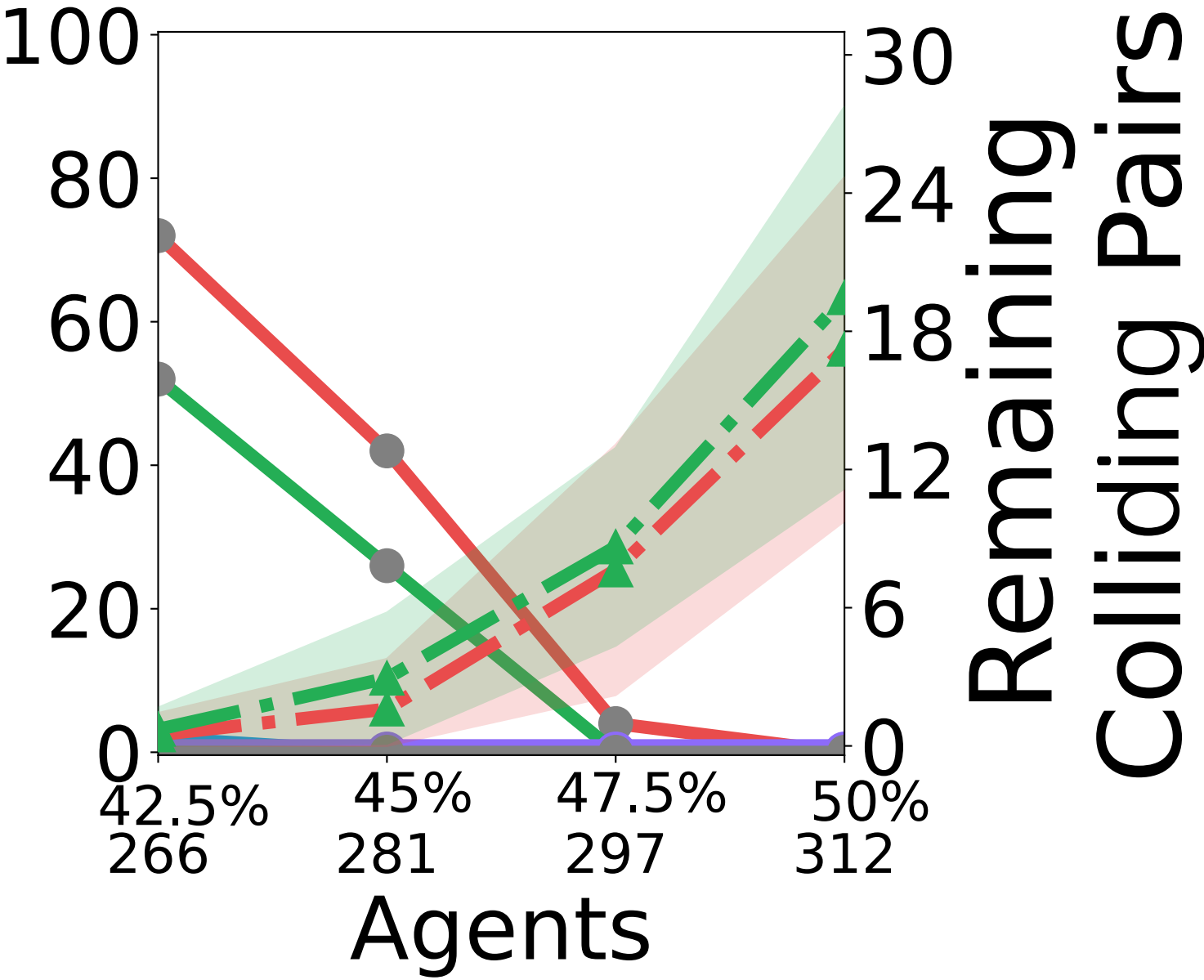}
    \caption{Warehouse}
    \label{fig:Warehouse}
\end{subfigure}
\begin{subfigure}[b]{0.18\textwidth}
    \includegraphics[width=\textwidth]{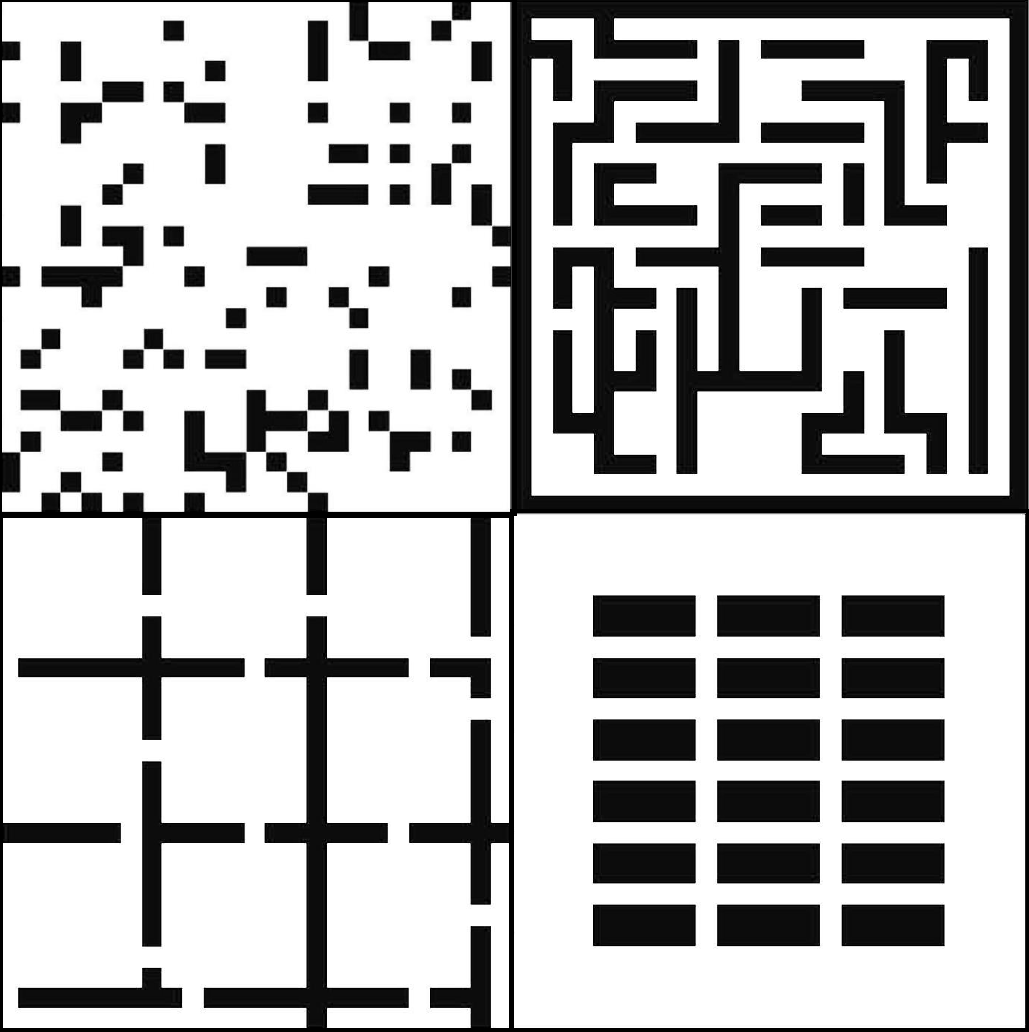}
    \caption{Map visualization}
    \label{fig:maps}
\end{subfigure}
\includegraphics[width=0.4\textwidth]{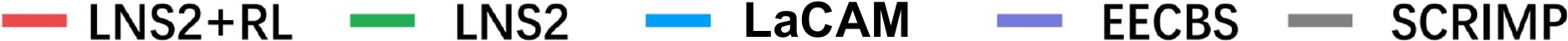}
\caption{Result on different map types and visualization of representative maps.
On the x-axis, the percentage represents agent density, while the number below it represents the number of agents.
The success rate is the percentage of tasks that are fully solved before the termination condition is met.
The remaining colliding pairs metric represents the number of CP remaining in the overall solution after reaching the time constraint and is only available for LNS2+RL and LNS2.
The shaded area represents one standard deviation of the remaining CP.
All three random maps have a 17.5\% obstacle density. 
Except for the random-small map with a size of $10 \times 10$ and the random-large map with a size of $50 \times 50$, all other five maps have a size of $25 \times 25$.
For maps with sizes $10 \times 10$, $25 \times 25$, and $50 \times 50$, the test counts are set to 100, 50, and 20, respectively, and the time constraints for each task, except when using SCRIMP, are set to 100 seconds, 600 seconds, and 5,000 seconds.
To ensure consistency with the original paper, SCRIMP uses the maximum timestep as the termination condition, with maximum timesteps of 356, 556, and 1,024 for maps sized $10 \times 10$, $25 \times 25$, and $50 \times 50$, respectively.
Figure~\ref{fig:maps} shows the visualization of the random-medium (top left), maze (top right), room (bottom left), and warehouse (bottom right) map.
}
\label{fig:all_map}
\end{figure*}

\subsubsection{Training Process}
\label{sec:train}
Our MARL model is trained by Proximal Policy Optimization (PPO)~\cite{schulman2017proximal}.
The total loss of the MARL model is $L_{final}=\omega_v \cdot L_v+\omega_{\pi} \cdot L_\pi +\omega_{invalid} \cdot L_{invalid} - \omega_e \cdot L_{e} $, where $L_v$ and $L_\pi$ are the standard critic and actor loss in PPO.
$L_{e}$ is the entropy of the agent's policy, suppressing its decline shown to encourage exploration and inhibit premature convergence.
$L_{invalid}$ is a binary cross-entropy loss (supervised invalid action loss) designed to reduce the logarithmic likelihood of selecting an invalid action. 
$\omega_v$, $\omega_{\pi}$, $\omega_{invalid}$, and $\omega_e$ are all manually set hyperparameters.
Details regarding the hyperparameters, computational resources, and training duration are provided in the supplementary materials.

The training of the MARL model is divided into two stages.
Both stages commence by generating a high-level MAPF task and obtaining the initial path set $P$ through PP+SIPPS.
Subsequently, a series of PMDO replanning tasks are generated to train the MARL model.
In the first training stage, to reduce training time, the MARL-controlled agents are randomly selected when constructing each PMDO task, bypassing adaptive neighborhood selection.
After a MARL episode ends, the path set $P$ is updated with the SIPPS paths $\hat{P}$ calculated at the beginning of the episode, regardless of whether these new paths reduce CP in $P$.
PMDO replanning tasks are continually generated and only switch to the next new MAPF task when the maximum iteration limit is reached.
Curriculum learning is adopted during the first stage of training, allowing agents to rapidly acquire fundamental skills from simpler tasks before progressively facing more challenging scenarios.
Specifically, we increase the task difficulty by increasing the hard obstacles rate and the total number of agents (MARL-controlled agents and soft obstacles) in the high-level MAPF tasks and advancing to greater difficulties once the predetermined volume of training data is accumulated.
As task difficulty increases, the advantage of SIPPS weakens, and the agents' preference for safely wandering in a certain area increases.
Therefore, we gradually lower the value of $\alpha$ in the off-route penalty to reduce references to SIPPS paths and gradually increase the action cost to incentivize agents to move toward their goal. 
Additionally, we employ imitation learning (behavior cloning based on the centralized planner ODrM*~\cite{wagner2015subdimensional}) at the beginning of each task difficulty in the first stage to explore high-quality regions of the agents' state-action space.
Upon reaching the manually predefined number of training steps, the training transitions to the second stage. 
This stage aims to acclimate the MARL policy to scenarios it will face when inserted into the LNS2+RL algorithm.
Therefore, both the selection of agents and the method for updating path set $P$ are exactly the same as the LNS2+RL algorithm's execution. 

\section{Experiments}
We compare LNS2+RL with vanilla LNS2\footnote{Code from 
\url{https://github.com/Jiaoyang-Li/MAPF-LNS2}} as well as a representative set of MAPF algorithms: the bounded-suboptimal algorithm EECBS (with a suboptimality factor $w = 5$)\footnote{Code from 
\url{https://github.com/Jiaoyang-Li/EECBS}}, the complete but unbounded-suboptimal algorithm LaCAM\footnote{Code from 
\url{https://github.com/Kei18/lacam}}, and the pure MARL-based algorithm SCRIMP\footnote{Code from 
\url{https://github.com/marmotlab/SCRIMP}}.
Code for LNS2, EECBS, LaCAM was written in C++, SCRIMP was written in Python, and LNS2+RL was written in both C++ and Python and connected using pybind. 
To obtain results faster, LNS2+RL, LNS2, EECBS used Ray to test multiple MAPF tasks simultaneously.
However, due to high memory demands (we limited the maximum memory to 100GB per task), LaCAM and SCRIMP were tested using a single process.
The same MARL model was used in all evaluations, and it was trained on maps with random obstacles.

\subsection{Comparison with Baselines}

The evaluation results in Figure~\ref{fig:all_map} show that LNS2+RL consistently achieves equal or higher SR compared to LNS2 across all tasks, except for those on the empty map with 77.5\% agent density (484 agents). 
It is noteworthy that the improvement of LNS2+RL on CP is more pronounced.
In tasks with 65\% agent density (65 agents) on the random-small map, despite both algorithms achieving the same SR, LNS2+RL remarkably reduces the remaining CP by 26.3\% compared to LNS2.
The structural complexity of maps significantly influences algorithm performance. 
Despite being trained on random maps without prior exposure to environments with complex obstacle structures, such as mazes, rooms, and warehouses, LNS2+RL demonstrates remarkable improvements in SR on these three types of maps.
Conversely, LaCAM excels on maps with simple or no obstacle structures, achieving a 100\% SR on empty maps, but close to 0\% SR on maps with complex obstacle structures.
Due to the high difficulty of the selected tasks, the SR of EECBS and SCRIMP remains at 0\% for most tasks.
Furthermore, taking the tasks in Table~\ref{table:SoC} as an example, the SoC for LNS2+RL only averages 1.03\% higher than that of LNS2.
Despite LaCAM displaying higher SR on the random-small map, its SoC notably lags behind.
Conversely, EECBS consistently achieves the best SoC but at the expense of obtaining the lower SR. 
SCRIMP exhibits the poorest performance in both SR and SoC.
Moreover, LNS2+RL and LNS2 offer the added benefit of maintaining consistent memory usage, in contrast to the other three baselines, whose memory usage escalates rapidly over time due to the generation of longer paths, larger search frontiers, and more collisions.
\begin{table}[h]
\centering
\scalebox{0.8}{
\begin{tabular}{|c|rrrr|}
\hline
\multirow{2}{*}{Algorithms} & \multicolumn{4}{c|}{Sum of Cost} \\ \cline{2-5} 
 & \multicolumn{1}{c|}{\begin{tabular}[c]{@{}c@{}}50\% agent \\density \end{tabular}} & \multicolumn{1}{c|}{\begin{tabular}[c]{@{}c@{}}55\% agent \\density \end{tabular}} & \multicolumn{1}{c|}{\begin{tabular}[c]{@{}c@{}}60\% agent \\density \end{tabular}} & \begin{tabular}[c]{@{}c@{}}65\% agent\\ density\end{tabular} \\ \hline
LNS2+RL & \multicolumn{1}{r|}{852.7} & \multicolumn{1}{r|}{942.6} & \multicolumn{1}{r|}{994.9} & 1,052.8 \\
LNS2 & \multicolumn{1}{r|}{831.0} & \multicolumn{1}{r|}{939.4} & \multicolumn{1}{r|}{972.4} & 1,064.2 \\
LaCAM & \multicolumn{1}{r|}{2,395.2} & \multicolumn{1}{r|}{5,288.3} & \multicolumn{1}{r|}{5,035.1} & 306,438.3 \\
EECBS & \multicolumn{1}{r|}{825.1} & \multicolumn{1}{r|}{912.7} & \multicolumn{1}{r|}{929.7} & 1,030.0 \\
SCRIMP & \multicolumn{1}{r|}{3,719.4} & \multicolumn{1}{r|}{4,444.3} & \multicolumn{1}{r|}{9,869.7} & - \\ \hline
\end{tabular}}
\caption{
Sum of costs on random-small map (accounting only for successful tasks).
”-” represents unavailable data due to 0\% SR.}
\label{table:SoC}
\end{table}

\subsection{Performance Analysis}
\begin{figure}[h]
  \centering
  \includegraphics[width=0.4\linewidth]{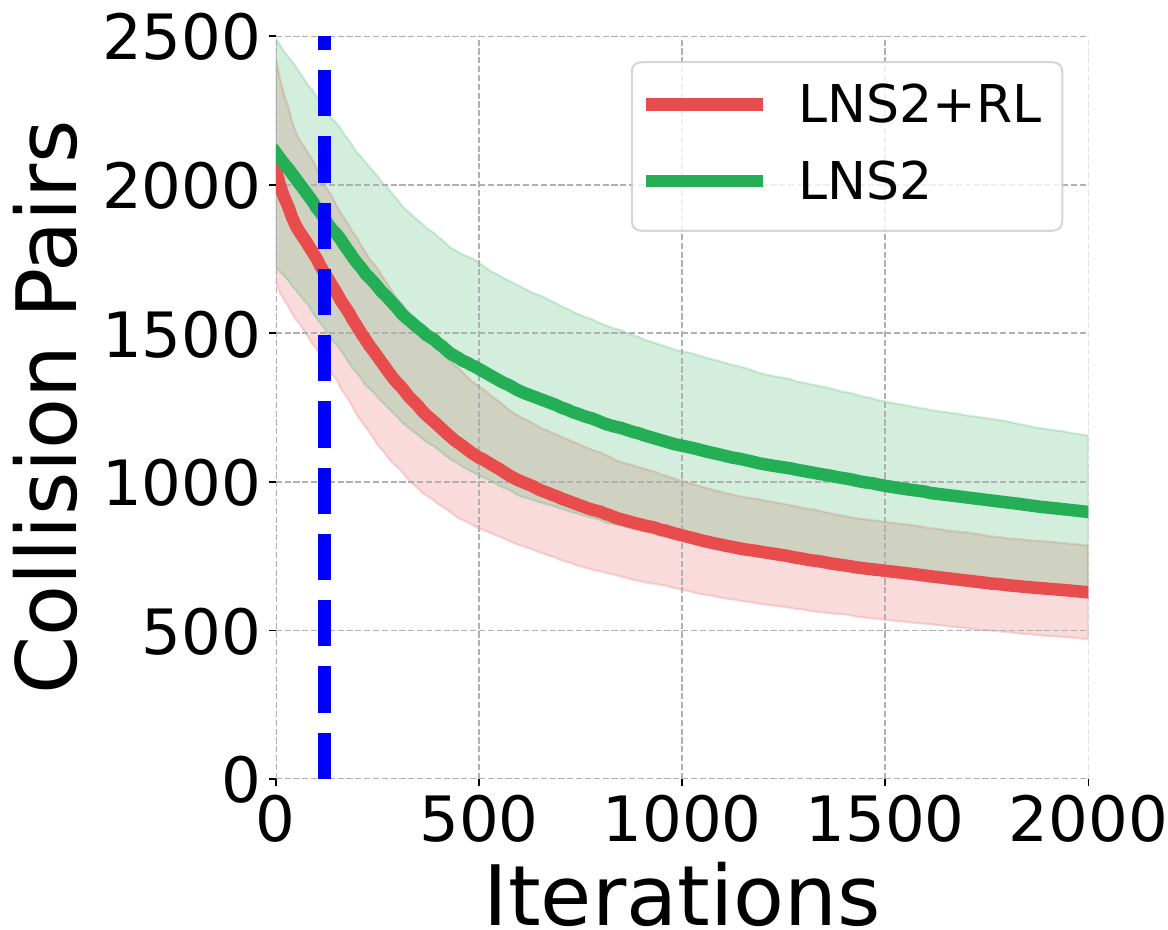}
  \caption{ 
Number of remaining CP through the iterations.
The dashed blue line shows the switch from MARL to PP+SIPPS.}
  \label{fig:curve}
\end{figure}

Table~\ref{table:single} shows the performance ratio of the solution obtained by the MARL policy compared to that of PP+SIPPS in replanning tasks across different numbers of iterations.
The MARL policy clearly produces fewer collisions overall, but its advantage in reducing CP is less pronounced, indicating that the MARL policy tends to cause collisions with different agents.
With replanning task difficulty decreasing as iterations increase, the MARL policy's advantage in reducing CP diminishes. 
Given that the speed of the MARL policy is nearly seven times slower than that of PP+SIPPS, switching to PP+SIPPS becomes a necessary trade-off between time and solution quality (the number of collisions in the solution).
Figure~\ref{fig:curve} illustrates the trend of the number of remaining CP over 2000 iterations in MAPF tasks on random-medium maps with a 65\% agent density.
As expected, in the early stage, before switching to PP+SIPPS, the number of CP for LNS2+RL decreases at a faster rate.
After switching, the gap between LNS2+RL and LNS2 continues to widen, as the high solution quality achieved by the MARL policy reduces the difficulty of subsequent replanning tasks.
\begin{table}[h] 
\centering
\scalebox{0.8}{
\begin{tabular}{|c|c|c|c|c|}
\hline
Iterations & Collisions & Colliding Pairs & Sum of Cost & Time    \\ \hline
1          & -73.7\%    & -14.6\%         & +35.0\%     & +1,175\% \\
50         & -72.5\%    & -1.2\%          & +19.7\%     & +737\%  \\
100        & -73.2\%    & +2.3\%          & +15.5\%     & +633\%  \\ \hline
\end{tabular}}
\caption{
Performance ratio of the solution obtained by the MARL policy compared to that of PP+SIPPS on low-level replanning tasks in random-medium maps with 65\% agent density.
The difference between collisions and CP is that for $n$ collisions involving the same agents, the number of collisions is counted as $n$, but CP is counted as $1$.}
\label{table:single}
\end{table}

\subsection{Real World Experiment}

We implemented our algorithm on a team of 90 virtual and 10 real robots in a warehouse environment.
The planned path may not be executed accurately due to disturbances and control inaccuracies.
To eliminate these errors, we implement an \textit{Action Dependency Graph} (ADG) \cite{honig2019persistent}.
From our experiment, we observe that agents can reach their goals quickly without collisions, and errors are eliminated by the ADG, demonstrating the potential of using our method in the real world.
More information about our experiment are included in the supplementary material.

\section{Conclusion}

This paper introduces LNS2+RL, a novel MAPF algorithm that combines MARL with large neighborhood search. It follows the overall framework of LNS2, beginning with a path set that may contain collisions and then iteratively replanning paths for subsets of agents until the path set becomes collision-free. 
In early iterations, LNS2+RL addresses these challenging replanning tasks with a MARL-based planner, which offers higher solution quality (fewer collisions) but at a slower pace.
As iterations progress, it adaptively switches to a fast, prioritized planning algorithm to quickly resolve remaining collisions.
Our experiments on high-agent-density tasks across various world sizes, team sizes, and map structures consistently demonstrate the superior performance of LNS2+RL compared to other state-of-the-art MAPF algorithms.
The practicality of LNS2+RL is validated in a hybrid warehouse mockup simulation.

\section{Acknowledgments}
This research was supported by the Singapore Ministry of Education (MOE), as well as by an Amazon Research Award.

\section{Additional Results}
\subsection{Detailed Results}
The detailed results of the Experiment Section in the main body of the paper can be found in Table~\ref{table:all_data}.
\begin{sidewaystable*}[]
\centering
\scalebox{0.8}{
\begin{tabular}{|c|cccccccccccccccc|}
\hline
\multirow{2}{*}{Methods} & \multicolumn{16}{c|}{Random-small: 10 * 10 world size with 17.5\% static obstacle rate and 50\%(50 agents), 55\%(55 agents), 60\%(60 agents), 65\%(65 agents) agents density} \\ \cline{2-17} 
 & \multicolumn{4}{c|}{Success Rate ↑} & \multicolumn{4}{c|}{Sum of Cost ↓} & \multicolumn{4}{c|}{Runtime ↓} & \multicolumn{4}{c|}{Remaining Colliding Pairs ↓} \\ \hline
LNS2+RL & \textbf{82\%} & \textbf{70\%} & 59\% & \multicolumn{1}{c|}{27\%} & 852.7 & 942.6 & 994.9 & \multicolumn{1}{c|}{1,052.8} & 21.6 & 34.9 & 49.8 & \multicolumn{1}{c|}{79.9} & \textbf{\begin{tabular}[c]{@{}c@{}}0.45\\ (-25.0\%)\end{tabular}} & \textbf{\begin{tabular}[c]{@{}c@{}}1.06\\ (-29.8\%)\end{tabular}} & \textbf{\begin{tabular}[c]{@{}c@{}}2.70\\ (-27.2\%)\end{tabular}} & \textbf{\begin{tabular}[c]{@{}c@{}}7.44\\ (-25.7\%)\end{tabular}} \\
LNS2 & 80\% & 68\% & 50\% & \multicolumn{1}{c|}{27\%} & 831.0 & 939.4 & 972.4 & \multicolumn{1}{c|}{1,064.2} & \textbf{20.8} & 33.9 & 54.1 & \multicolumn{1}{c|}{77.0} & 0.60 & 1.51 & 3.71 & 10.01 \\
EECBS & 38\% & 18\% & 3\% & \multicolumn{1}{c|}{1\%} & \textbf{825.1} & \textbf{912.7} & \textbf{929.7} & \multicolumn{1}{c|}{\textbf{1,030.0}} & 63.9 & 82.8 & 97.1 & \multicolumn{1}{c|}{99.1} & - & - & - & - \\
LaCAM & 75\% & \textbf{70\%} & \textbf{68\%} & \multicolumn{1}{c|}{\textbf{61\%}} & 2,395.2 & 5,288.3 & 5,035.1 & \multicolumn{1}{c|}{306,438.3} & 27.0 & \textbf{32.2} & \textbf{34.8} & \multicolumn{1}{c|}{\textbf{42.4}} & - & - & - & - \\
SCRIMP & 22\% & 12\%& 3\% & \multicolumn{1}{c|}{0\%} & 3,719.4 & 4,444.3 & 9,869.7 & \multicolumn{1}{c|}{-} & 2.7 & 3.3 & 4.2 & \multicolumn{1}{c|}{5.1} & - & - & - & - \\\hline

 & \multicolumn{16}{c|}{Random-medium: 25 * 25 world size with 17.5\% static obstacle rate and 50\%(312 agents), 55\%(344 agents), 60\%(375 agents), 65\%(406 agents) agents density} \\ \hline
LNS2+RL & \textbf{72\%} & \textbf{30\%} & 2\% & \multicolumn{1}{c|}{0\%} & 16,726.3 & 16,727.7 & 16,605.0 & \multicolumn{1}{c|}{-} & \textbf{222.1} & \textbf{486.8} & 593.5 & \multicolumn{1}{c|}{600.0} & \textbf{\begin{tabular}[c]{@{}c@{}}0.80\\ (-20.0\%)\end{tabular}} & \textbf{\begin{tabular}[c]{@{}c@{}}3.54\\ (-53.2\%)\end{tabular}} & \textbf{\begin{tabular}[c]{@{}c@{}}32.52\\ (-46.5\%)\end{tabular}} & \textbf{\begin{tabular}[c]{@{}c@{}}165.38\\ (-34.0\%)\end{tabular}} \\
LNS2 & 70\% & 20\% & 2\% & \multicolumn{1}{c|}{0\%} & \textbf{14,157.5} & \textbf{15,037.1} & \textbf{16,513.0} & \multicolumn{1}{c|}{-} & 256.8 & 509.8 & \textbf{591.7} & \multicolumn{1}{c|}{600.0} & 1.00 & 7.56 & 60.82 & 250.66 \\
EECBS & 0\% & 0\% & 0\% & \multicolumn{1}{c|}{0\%} & - & - & - & \multicolumn{1}{c|}{-} & 600.0 & 600.0 & 600.0 & \multicolumn{1}{c|}{600.0} & - & - & - & - \\
LaCAM & 36\% & 26\% & \textbf{28\%} & \multicolumn{1}{c|}{\textbf{18\%}} & 35,070.2 & 70,508.0 & 66,149.3 & \multicolumn{1}{c|}{\textbf{127,656.0}} & 397.4 & - & - & \multicolumn{1}{c|}{-} & - & - & - & - \\ 
SCRIMP & 0\% & 0\%& 0\% & \multicolumn{1}{c|}{0\%} & - & - & - & \multicolumn{1}{c|}{-} & 96.8 & 140.9 & 210.2  & \multicolumn{1}{c|}{284.5} & - & - & - & - \\ \hline

 & \multicolumn{16}{c|}{Random-large: 50 * 50 world size with 17.5\% static obstacle rate and 50\%(1,250 agents), 55\%(1,375 agents), 60\%(1,500 agents), 65\%(1,625 agents) agents density} \\ \hline
LNS2+RL & \textbf{35\%} & 0\% & 0\% & \multicolumn{1}{c|}{0\%} & \textbf{127,514.3} & - & - & \multicolumn{1}{c|}{-} & \textbf{4,308.7} & 5,000.0 & 5,000.0 & \multicolumn{1}{c|}{5,000.0} & \textbf{\begin{tabular}[c]{@{}c@{}}2.95\\ (-42.7\%)\end{tabular}} & \textbf{\begin{tabular}[c]{@{}c@{}}108.30\\ (-21.2\%)\end{tabular}} & \textbf{\begin{tabular}[c]{@{}c@{}}968.85\\ (-35.1\%)\end{tabular}} & \textbf{\begin{tabular}[c]{@{}c@{}}3,484.00\\ (-18.3\%)\end{tabular}} \\
LNS2 & 30\% & 0\% & 0\% & \multicolumn{1}{c|}{0\%} & 130,756.5 & - & - & \multicolumn{1}{c|}{-} & 4,329.2 & 5,000.0 & 5,000.0 & \multicolumn{1}{c|}{5,000.0} & 5.15 & 137.45 & 1,477.10 & 4,261.85 \\
EECBS & 0\% & 0\% & 0\% & \multicolumn{1}{c|}{0\%} & - & - & - & \multicolumn{1}{c|}{-} & 5,000.0 & 5,000.0 & 5,000.0 & \multicolumn{1}{c|}{5,000.0} & - & - & - & - \\
LaCAM & 5\% & \textbf{25\%} & 0\% & \multicolumn{1}{c|}{\textbf{5\%}} & 204,820.0 & \textbf{955,662.6} & - & \multicolumn{1}{c|}{\textbf{474,872.0}} & - & - & - & \multicolumn{1}{c|}{-} & - & - & - & - \\
SCRIMP & 0\% & 0\%& 0\% & \multicolumn{1}{c|}{0\%} & - & - & - & \multicolumn{1}{c|}{-} & - & -  & - & \multicolumn{1}{c|}{-} & - & - & - & - \\ \hline

 & \multicolumn{16}{c|}{Empty: 25 * 25 world size with 0\% static obstacle rate and 72.5\%(453 agents), 75\%(469 agents), 77.5\%(484 agents), 80\%(500 agents) agents density} \\ \hline
LNS2+RL & \textbf{100\%} & 84\% & 6\% & \multicolumn{1}{c|}{0\%} & 20,155.2 & 20,132.5 & \textbf{19,278.0} & \multicolumn{1}{c|}{-} & 94.7 & 333.5 & 596.6 & \multicolumn{1}{c|}{600.0} & \textbf{\begin{tabular}[c]{@{}c@{}}0.00\\ (-0.0\%)\end{tabular}} & \textbf{\begin{tabular}[c]{@{}c@{}}0.26\\ (-66.7\%)\end{tabular}} & \textbf{\begin{tabular}[c]{@{}c@{}}4.04\\ (-1.9\%)\end{tabular}} & \textbf{\begin{tabular}[c]{@{}c@{}}19.08\\ (-11.0\%)\end{tabular}} \\
LNS2 & \textbf{100\%} & 68\% & 18\% & \multicolumn{1}{c|}{0\%} & \textbf{19,365.8} & \textbf{19,485.4} & 19,390.4 & \multicolumn{1}{c|}{-} & 117.7 & 403.8 & 575.5 & \multicolumn{1}{c|}{600.0} & \textbf{0.00} & 0.78 & 4.12 & 21.44 \\
EECBS & 0\% & 0\% & 0\% & \multicolumn{1}{c|}{0\%} & - & - & - & \multicolumn{1}{c|}{-} & 600.0 & 600.0 & 600.0 & \multicolumn{1}{c|}{600.0} & - & - & - & - \\
LaCAM & \textbf{100\%} & \textbf{100\%} & \textbf{100\%} & \multicolumn{1}{c|}{\textbf{100\%}} & 24,834.3 & 26,929.9 & 29,408.2 & \multicolumn{1}{c|}{\textbf{32,298.9}} & \textbf{0.006} & \textbf{0.007} & \textbf{0.007} & \multicolumn{1}{c|}{\textbf{0.008}} & - & - & - & - \\ 
 SCRIMP & 0\% & 0\%& 0\% & \multicolumn{1}{c|}{0\%} & - & - & - & \multicolumn{1}{c|}{-} & 167.8 & 217.3 & 286.4 & \multicolumn{1}{c|}{346.5} & - & - & - & - \\ \hline

 & \multicolumn{16}{c|}{Maze: 25 * 25 world size with 45.76\% static obstacle rate and 20\%(125 agents), 22.5\%(141 agents), 25\%(156 agents), 27.5\%(172 agents) agents density} \\ \hline
LNS2+RL & \textbf{100\%} & \textbf{78\%} & \textbf{44\%} & \multicolumn{1}{c|}{\textbf{6\%}} & 5,724.3 & 6,523.3 & 7,307.6 & \multicolumn{1}{c|}{\textbf{8,698.0}} & \textbf{16.4} & \textbf{183.6} & \textbf{416.0} & \multicolumn{1}{c|}{\textbf{568.8}} & \textbf{\begin{tabular}[c]{@{}c@{}}0.00\\ (-100\%)\end{tabular}} & \textbf{\begin{tabular}[c]{@{}c@{}}0.34\\ (-10.5\%)\end{tabular}} & \textbf{\begin{tabular}[c]{@{}c@{}}1.24\\ (-51.6\%)\end{tabular}} & \textbf{\begin{tabular}[c]{@{}c@{}}4.88\\ (-50.8\%)\end{tabular}} \\
LNS2 & 92\% & 74\% & 26\% & \multicolumn{1}{c|}{0\%} & \textbf{5,563.3} & \textbf{6,205.6} & \textbf{7,210.6} & \multicolumn{1}{c|}{-} & 67.6 & 217.9 & 493.5 & \multicolumn{1}{c|}{600.0} & 0.08 & 0.38 & 2.56 & 9.92 \\
EECBS & 6\% & 0\% & 0\% & \multicolumn{1}{c|}{0\%} & - & - & - & \multicolumn{1}{c|}{-} & 600.0 & 600.0 & 600.0 & \multicolumn{1}{c|}{600.0} & - & - & - & - \\
LaCAM & 2\% & 0\% & 0\% & \multicolumn{1}{c|}{0\%} & 41,924.0 & - & - & \multicolumn{1}{c|}{-} & 588.8 & 600.0 & 600.0 & \multicolumn{1}{c|}{-} & - & - & - & - \\ 
 SCRIMP & 0\% & 0\%& 0\% & \multicolumn{1}{c|}{0\%} & - & - & - & \multicolumn{1}{c|}{-} & 14.1 & 19.2 & 25.2 & \multicolumn{1}{c|}{33.6} & - & - & - & - \\ \hline

 & \multicolumn{16}{c|}{Room: 25 * 25 world size with 19.52\% static obstacle rate and 40\%(250 agents), 45\%(281 agents), 50\%(321 agents), 55\%(344 agents) agents density} \\ \hline
LNS2+RL & \textbf{100\%} & \textbf{98\%} & \textbf{12\%} & \multicolumn{1}{c|}{0\%} & 35,497.9 & 44,779.1 & 58,620.2 & \multicolumn{1}{c|}{-} & \textbf{41.3} & \textbf{154.5} & \textbf{577.9} & \multicolumn{1}{c|}{600.0} & \textbf{\begin{tabular}[c]{@{}c@{}}0.00\\ (-100\%)\end{tabular}} & \textbf{\begin{tabular}[c]{@{}c@{}}0.02\\ (-99.9\%)\end{tabular}} & \textbf{\begin{tabular}[c]{@{}c@{}}12.14\\ (-91.8\%)\end{tabular}} & \textbf{\begin{tabular}[c]{@{}c@{}}179.08\\ (-69.3\%)\end{tabular}} \\
LNS2 & 90\% & 36\% & 2\% & \multicolumn{1}{c|}{0\%} & \textbf{26,660.6} & \textbf{35,221.3} & \textbf{39,378.0} & \multicolumn{1}{c|}{-} & 167.1 & 467.9 & 599.2 & \multicolumn{1}{c|}{600.0} & 0.34 & 19.44 & 148.22 & 582.5 \\
EECBS & 0\% & 0\% & 0\% & \multicolumn{1}{c|}{0\%} & - & - & - & \multicolumn{1}{c|}{-} & 600.0 & 600.0 & 600.0 & \multicolumn{1}{c|}{600.0} & - & - & - & - \\
LaCAM & 0\% & 0\% & 0\% & \multicolumn{1}{c|}{0\%} & - & - & - & \multicolumn{1}{c|}{-} & 600.0 & 600.0 & 600.0 & \multicolumn{1}{c|}{600.0} & - & - & - & - \\ 
SCRIMP & 0\% & 0\%& 0\% & \multicolumn{1}{c|}{0\%} & - & - & - & \multicolumn{1}{c|}{-} & 85.8 & 123.4 & 157.1& \multicolumn{1}{c|}{200.9} & - & - & - & - \\ \hline

 & \multicolumn{16}{c|}{Warehouse: 25 * 25 world size with 28.8\% static obstacle rate and 42.5\%(266 agents), 45\%(281 agents), 47.5\%(297 agents), 50\%(312 agents) agents density} \\ \hline
LNS2+RL & \textbf{72\%} & \textbf{42\%} & \textbf{4\%} & \multicolumn{1}{c|}{0\%} & 11,202.9 & 12,240.2 & \textbf{12,082.0} & \multicolumn{1}{c|}{-} & \textbf{296.9} & \textbf{458.9} & \textbf{595.6} & \multicolumn{1}{c|}{600.0} & \textbf{\begin{tabular}[c]{@{}c@{}}0.48\\ (-33.3\%)\end{tabular}} & \textbf{\begin{tabular}[c]{@{}c@{}}1.60\\ (-45.6\%)\end{tabular}} & \textbf{\begin{tabular}[c]{@{}c@{}}6.58\\ (-23.8\%)\end{tabular}} & \textbf{\begin{tabular}[c]{@{}c@{}}17.22\\ (-11.5\%)\end{tabular}} \\
LNS2 & 52\% & 26\% & 0\% & \multicolumn{1}{c|}{0\%} & \textbf{11,027.3} & \textbf{11,892.8} & - & \multicolumn{1}{c|}{-} & 383.2 & 526.0 & 600.0 & \multicolumn{1}{c|}{600.0} & 0.72 & 2.94 & 8.64 & 19.46 \\
EECBS & 0\% & 0\% & 0\% & \multicolumn{1}{c|}{0\%} & - & - & - & \multicolumn{1}{c|}{-} & 600.0 & 600.0 & 600.0 & \multicolumn{1}{c|}{600.0} & - & - & - & - \\
LaCAM & 2\% & 0\% & 0\% & \multicolumn{1}{c|}{0\%} & 3,061,736.0 & - & - & \multicolumn{1}{c|}{-} & - & - & - & \multicolumn{1}{c|}{-} & - & - & - & - \\ 
 SCRIMP & 0\% & 0\%& 0\% & \multicolumn{1}{c|}{0\%} & - & - & - & \multicolumn{1}{c|}{-} & 60.8 & 73.2 & 92.3 & \multicolumn{1}{c|}{114.6} & - & - & - & - \\ \hline
\end{tabular}}
\caption{Average performance along MAPF metrics of different algorithms for instances with different world sizes, team sizes, and map structure.
For maps with sizes $10 \times 10$, $25 \times 25$, and $50 \times 50$, the test counts are set to 100, 50, and 20, respectively, and the time constraints for each task, except when using SCRIMP, are set to 100 seconds, 600 seconds, and 5,000 seconds.
To ensure consistency with the original paper, SCRIMP uses the maximum timestep as the termination condition, with maximum timesteps of 356, 556, and 1024 for maps sized $10 \times 10$, $25 \times 25$, and $50 \times 50$, respectively.
The success rate is the percentage of tasks fully solved before the termination condition is met(i.e., where all agents reached their goal without collisions).
The sum of cost only accounts for successful tasks.
Partial runtime data for LaCAM and SCRIMP is unavailable because some tasks in the test exceeded the maximum memory limit we set (100GB), leading to premature termination.
The remaining colliding pairs metric represents the number of CP remaining in the overall solution after reaching the time constraint and is only available for LNS2+RL and LNS2.
Percentages in parentheses represent the reduced remaining CP ratio of LNS2+RL compared to LNS2.
↑ indicates that “higher is better”, and ↓ “lower is better”. ”-” represents unavailable data.
The best-performing algorithms in each task are highlighted in bold. 
Because SCRIMP uses maximum timestep as the termination condition, it is excluded from the selection of the shortest time.}
\label{table:all_data}
\end{sidewaystable*}

\subsection{Results for Different Time Limits and Easier Tasks}
\begin{figure*}[h]
\captionsetup[subfigure]{aboveskip=-1pt} 
\centering
\begin{subfigure}[b]{0.23\textwidth}
    \includegraphics[width=\textwidth]{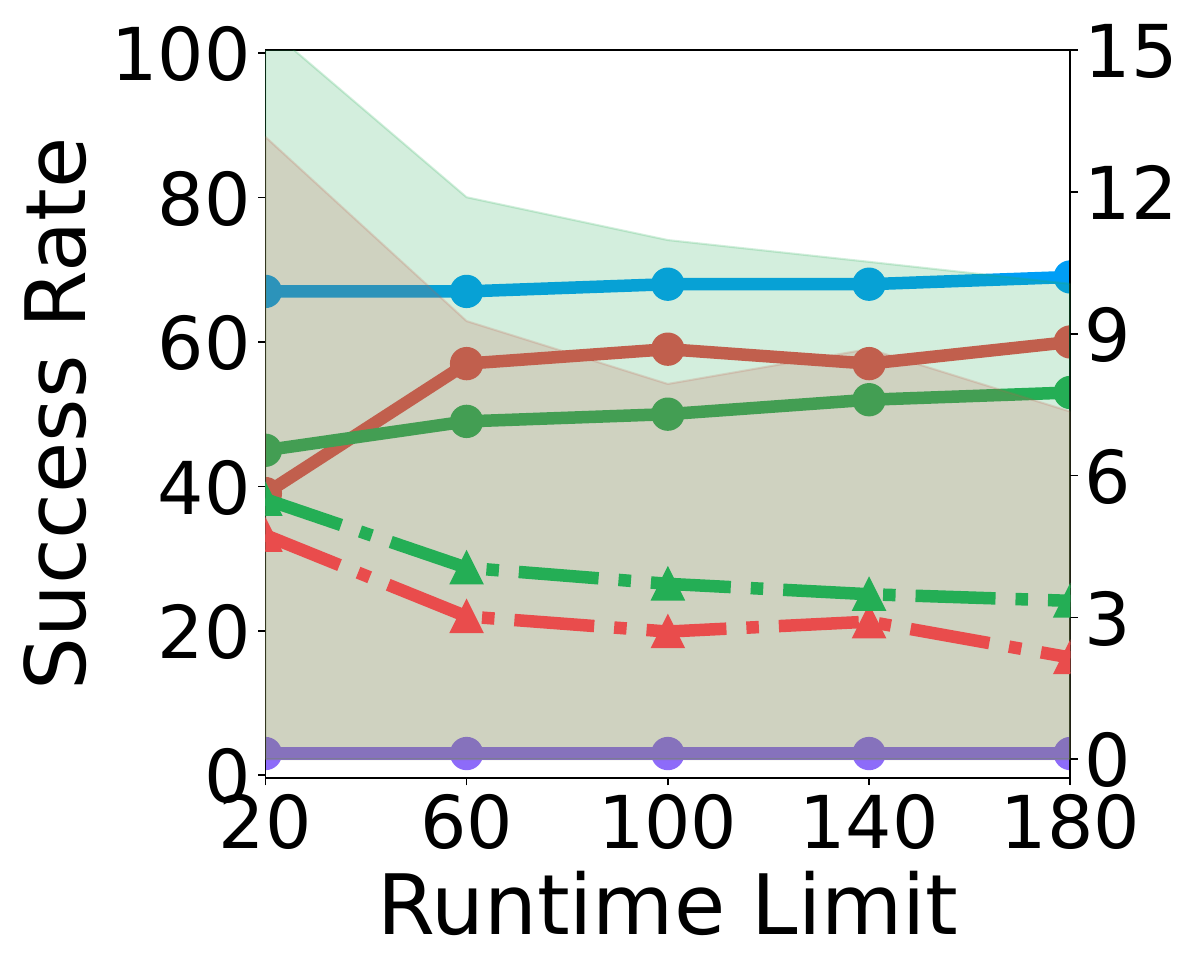}
    \caption{Result on the random-small map with 60\% agent density and different runtime limit.}
    \label{fig:diff_random}
\end{subfigure}
\begin{subfigure}[b]{0.23\textwidth}
    \includegraphics[width=\textwidth]{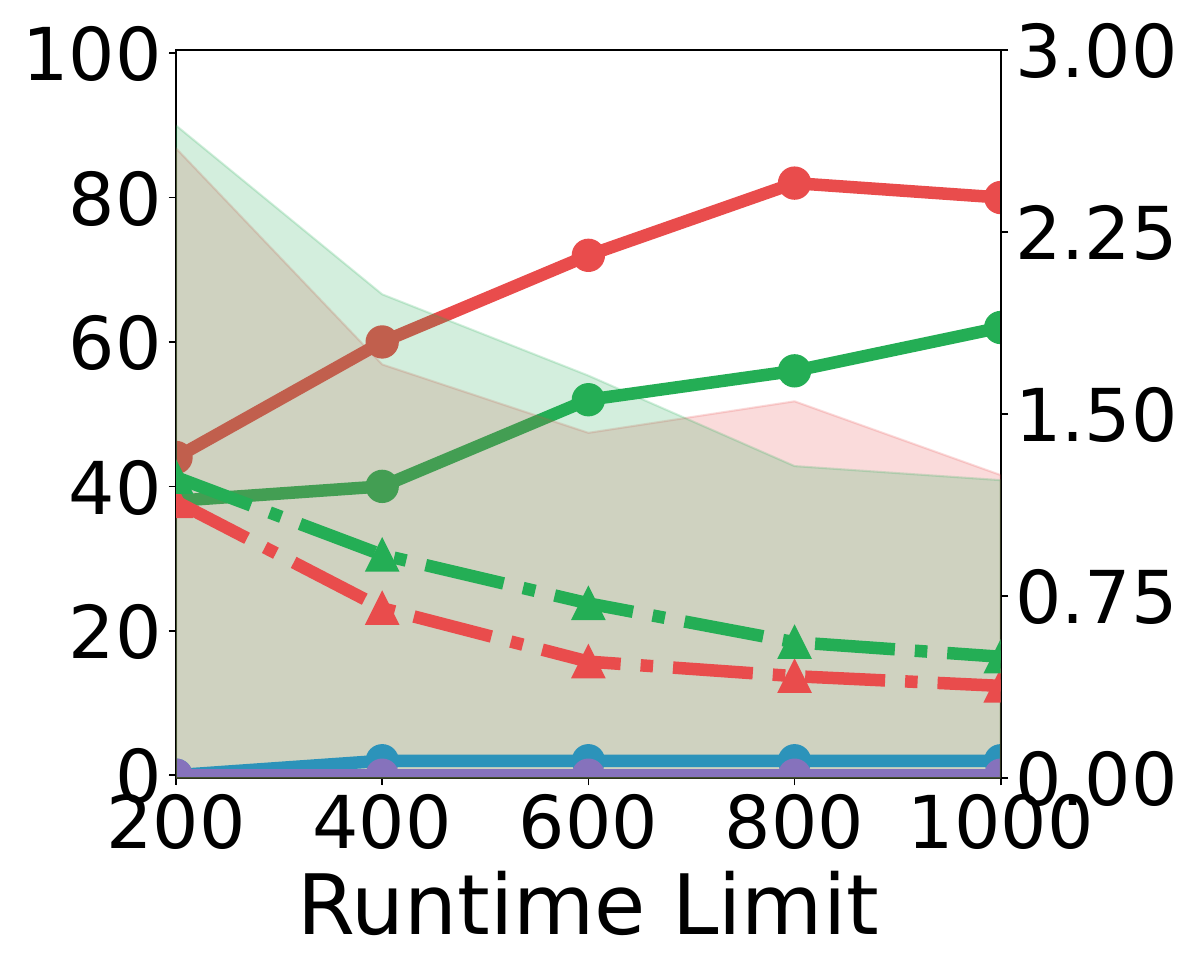}
    \caption{Result on the warehouse map with 42.5\% agent density and different runtime limit.}
    \label{fig:diff_warehouse}
\end{subfigure}
\begin{subfigure}[b]{0.23\textwidth}
    \includegraphics[width=\textwidth]{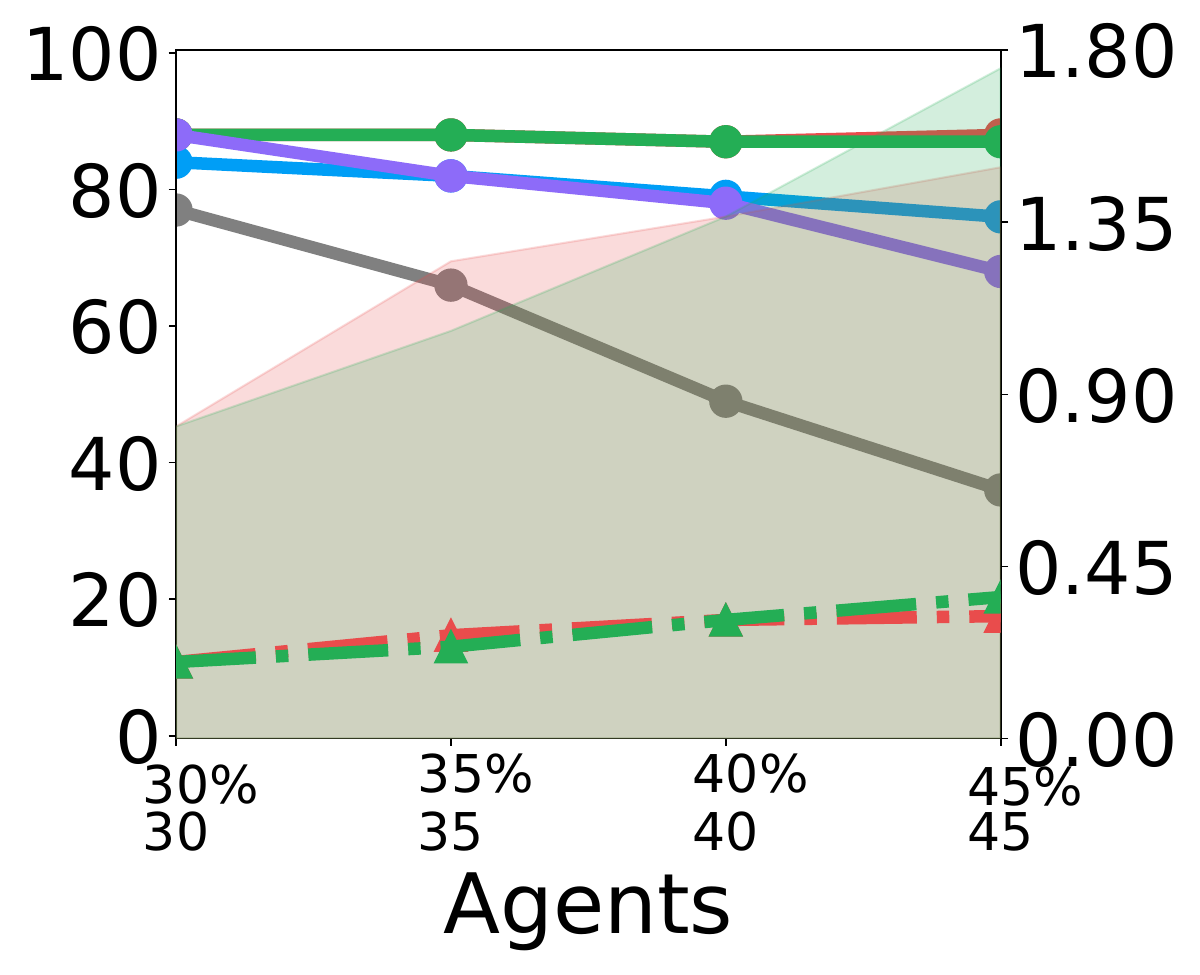}
    \caption{Result on the random-small map with 100 second runtime limit and different agent density.}
    \label{fig:easy_random}
\end{subfigure}
\begin{subfigure}[b]{0.23\textwidth}
    \includegraphics[width=\textwidth]{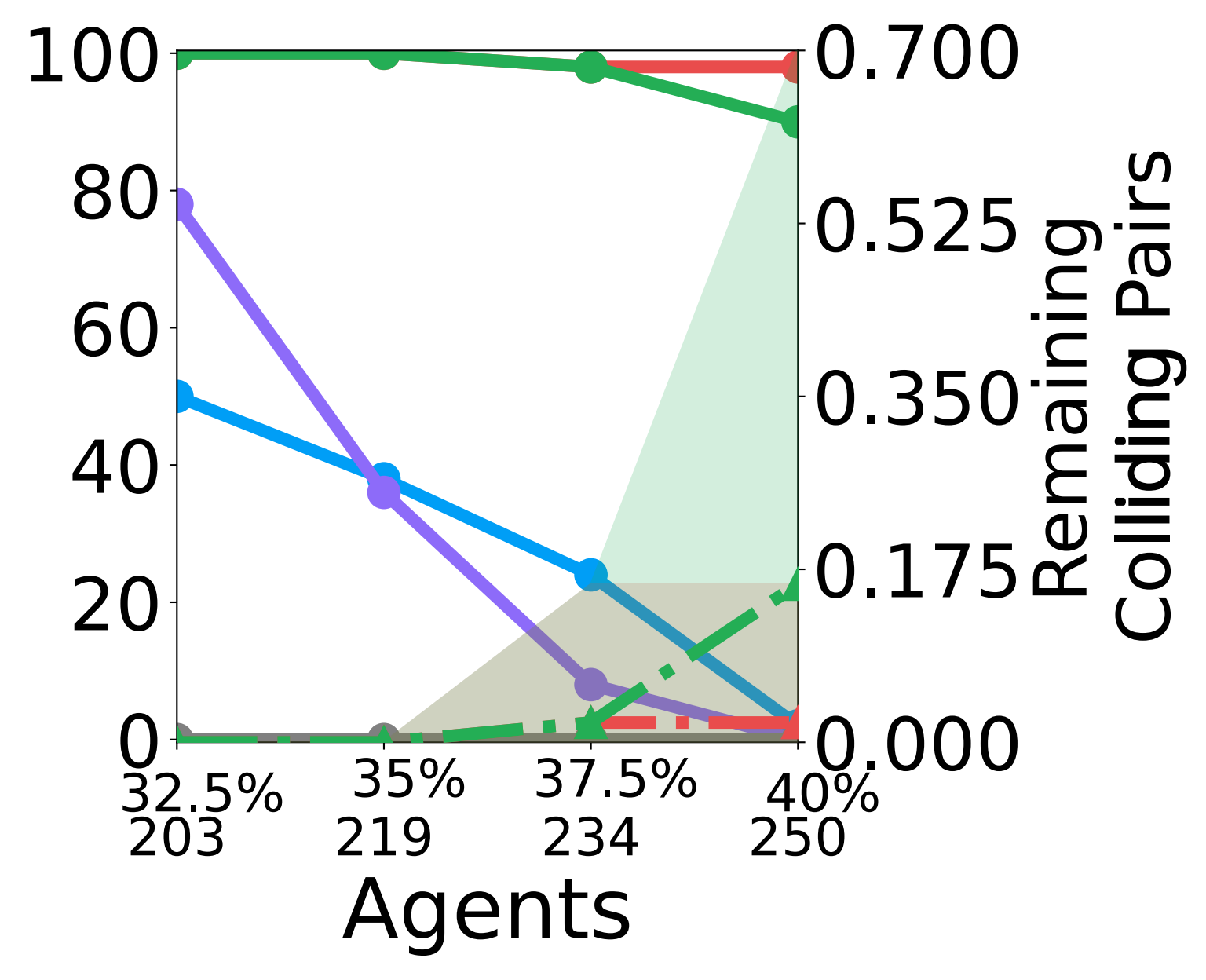}
    \caption{Result on the warehouse map with 600 second runtime limit and different agent density.}
    \label{fig:easy_warehouse}
\end{subfigure}
\includegraphics[width=0.45\textwidth]{label.pdf}
\caption{Result with different runtime limit and lower agent density.}
\label{fig:app_map}
\end{figure*}
In this section, we use the random-small map and warehouse map as representatives of maps with simple obstacle structure and maps with complex obstacle structure, respectively, to evaluate the performance of algorithms under varying runtime limits and task difficulties.
SCRIMP is not included in the runtime-limit experiment due to its different termination condition.
Specifically, we conducted the runtime-limit experiments on the random-small map with a 60\% agent density and the warehouse map with a 42.5\% agent density.
Their results are shown in Figure~\ref{fig:diff_random} and~\ref{fig:diff_warehouse}.
These results indicate that: First, the relative performance ranking of the algorithms remains consistent across different runtime limits, as discussed in the main paper.
LNS2+RL maintains its superiority over LNS2 in both SR and CP.
Second, each algorithm has a distinct runtime limit threshold, beyond which its SR stabilizes.
Notably, the thresholds of LNS2+RL and LNS2 exceed those of LaCAM and EECBS.
For instance, as depicted in Figure~\ref{fig:diff_random}, increasing the runtime limit from 20 to 60 seconds results in an 18\% increase in SR for LNS2+RL and a 4\% increase for LNS2, while the SRs for LaCAM and EECBS remain unchanged.

Subsequently, we maintained a constant runtime limit/maximum timestep for SCRIMP while progressively reducing task difficulty (agent density).
The results, illustrated in Figures~\ref{fig:easy_random} and~\ref{fig:easy_warehouse}, indicate that: First, EECBS and LaCAM perform relatively well on less challenging tasks, but the performance gaps between them and LNS2+RL and LNS2 widen as task difficulty escalates.
SHRIMP still performs the worst across all task difficulties.
Second, although in simpler tasks, the less pronounced advantages of the MARL-planner over PP+SIPPS lead to an earlier meeting of the switching condition, thus reducing the proportion of MARL-planner usage.
However, LNS2+RL continues to achieve performance similar to that of LNS2.

\subsection{Impact of Different Hyperparameters}

Here, we take tasks on the random-small map with 60\% agent density as an example to show the impact of the hyperparameters used to insert the MARL planner into the overall framework.
These hyperparameters include the neighborhood size in MARL planning, the threshold $\rho$ to switch the replanning solver from MARL planner to PP+SIPPS, the time step $t_l$ to stop MARL planning, and the maximum length of the hybrid path $t_h$.
The results are shown in Table~\ref{table:neighbor}~\ref{table:threshold}~\ref{table:stop}~\ref{table:max}.
The hyperparameter values we selected outperformed others, yet the changes in hyperparameters did not exhibit a clear trend in SR. 
It's worth noting that the optimal values for hyperparameters may vary across different tasks, necessitating fine-tuning to achieve the best performance of LNS2+RL.
For instance, in handling challenging tasks, using larger neighborhood sizes and smaller switching thresholds can increase the proportion of paths computed by the MARL-planner, thereby further leveraging its advantages to improve SR.

\begin{table}[H]
\centering
\begin{tabular}{|c|c|c|c|}
\hline
Neighborhood Size & 4 & 8 & 16 \\ \hline
Success Rate & 55\% & \textbf{59\%} & 57\% \\ \hline
\end{tabular}
\caption{Success rate for different neighborhood sizes}
\label{table:neighbor}
\end{table}

\begin{table}[H]
\centering
\begin{tabular}{|c|c|c|c|c|c|}
\hline
Switch Threshold $\rho$ & 0.2 & 0.25 & 0.3 & 0.35 & 0.4 \\ \hline
Success Rate & 58\% & 57\% & \textbf{59\%} & 50\% & 56\% \\ \hline
\end{tabular}
\caption{Success rate for different $\rho$}
\label{table:threshold}
\end{table}

\begin{table}[H]
\centering
\begin{tabular}{|c|c|c|c|}
\hline
Stop MARL planning $t_l$ & 0.8 & 1.2 & 1.6 \\ \hline
Success Rate & 55\% & \textbf{59\%} & 58\% \\ \hline
\end{tabular}
\caption{Success rate for different $t_l$}
\label{table:stop}
\end{table}

\begin{table}[H]
\centering
\begin{tabular}{|c|c|c|c|}
\hline
Maximum Length $t_h$ & 1.8 & 2.2 & 2.6 \\ \hline
Success Rate & 57\% & \textbf{59\%} & 54\% \\ \hline
\end{tabular}
\caption{Success rate for different $t_h$}
\label{table:max}
\end{table}

\subsubsection{The Importance of Using SIPPS+PP for Path Supplementation}
\begin{table*}[]
\centering
\begin{tabular}{|c|cccc|cccc|}
\hline
\multirow{2}{*}{Iterations} & \multicolumn{4}{c|}{Number of Agents Reached Goal} & \multicolumn{4}{c|}{One-Shot Success Rate} \\ \cline{2-9} 
 & $t_l$=0.8 & $t_l$=1 & $t_l$=1.2 & $t_l$=1.4 & $t_l$=0.8 & $t_l$=1 & $t_l$=1.2 & $t_l$=1.4 \\ \hline
1 & 6.12 & 7.12 & 7.50 & 7.59 & 25\% & 64\% & 78\% & 80\% \\
50 & 6.01 & 6.96 & 7.37 & 7.47 & 22\% & 51\% & 69\% & 76\% \\
100 & 6.01 & 6.89 & 7.31 & 7.48 & 23\% & 51\% & 69\% & 76\% \\ \hline
\end{tabular}
\caption{Goal-reaching agents vary with number of iterations and $t_l$}
\label{table:reach}
\end{table*}
Since the MARL-based planner cannot guarantee solution completeness, LNS2+RL employs SIPPS+PP to supplement the remaining paths for agents that have not reached their goals by timestep $t_l$, thereby fully leveraging the MARL-planner's capabilities.
In this section, we present how the PMDO task one-shot success rate of our MARL-based planner varies with the number of iterations and $t_l$ on random-small maps with a 60\% agent ratio and a neighborhood size of 8, to demonstrate the importance of using SIPPS+PP for path supplementation.
As shown in Table~\ref{table:reach}, the success rate (the proportion of PMDO tasks where all agents reach their goals) of our MARL-based planner at $t_l=1.2$ and the first iteration is 78\%.
As the number of iterations increases, the task difficulty gradually rises, leading to a decline in the success rate. 
The number of agents reaching their goal per task remains around 7.5, indicating that the task failures are caused by a small number of agents failing to find paths.
This observation suggests that updating the overall path set solely with solutions where all agents reach their goals could result in the loss of many high-quality solutions already obtained. 
Additionally, we note that the benefits of increasing $t_l$ diminish over time.
For example, at the 50th iteration, increasing $t_l$ from 1 to 1.2 raises the average number of goal-reaching agents by 0.41, whereas increasing it from 1.2 to 1.4 only yields an additional 0.1 agents.
Therefore, we set $t_l=1.2$ to balance planning time and gains.

\subsection{Number of Iterations}
\begin{table}[]
\centering
\begin{tabular}{|c|c|c|c|c|}
\hline
Algorithms & 50 agents & 55 agents & 60 agents & 65 agents \\ \hline
LNS2+RL    & 364,549.8                & 377,825.4                & 410,457.8                & 407,344.2                \\
LNS2       & 373,025.6                & 382,657.5                & 451,732.3                & 453,540.0                \\ \hline
\end{tabular}
\caption{Average number of iterations on random-small maps with different numbers of agents.}
\label{table:iter}
\end{table}
Table~\ref{table:iter} shows the average number of iterations spent by LNS2+RL and LNS on random-small maps with different numbers of agents.
The results indicate that, compared to LNS, LNS2+RL requires fewer iterations to solve MAPF tasks.

\section{Hyperparameters}
\begin{table*}[h]
\centering
\scalebox{0.9}{
\begin{tabular}{l|l}
\hline
Hyperparameter & Value \\ \hline \hline
Number of MARL controlled agents & 8 \\ \hline
Ratio of total number of agents to world size & [(40\%, 45\%, 50\%), (50\%, 55\%, 60\%), (60\%, 65\%, 70\%)] \\ \hline
World size & (10, 25, 50) \\ \hline
Obstacle density & [(5\%, 7.5\%, 10\%), (10\%, 12.5\%, 15\%), (15\%, 17.5\%, 20\%)] \\ \hline
FOV size & 9 \\ \hline
Maximum episode length & 356 \\ \hline
$\alpha$ & (0.06, 0.05, 0.04)\\ \hline
Reward shaping coefficient & 0.2 \\ \hline
$\delta_c, \delta_d$ & 0.225, 0.075 \\ \hline
Spatial and temporal coefficient for trajectory prediction & 0.9, 0.1  \\ \hline
Curriculum learning switching difficulty timestep & (1e7, 2e7) \\ \hline

Learning rate & 1e-5 \\ \hline
Discount factor & 0.95 \\ \hline
Gae lamda & 0.95 \\ \hline
Clip parameter for probability ratio & 0.2 \\ \hline
Gradient clip norm & 50 \\ \hline
Entropy coefficient $\omega_e$ & 0.01 \\ \hline
Actor loss coefficient $\omega_{\pi}$ & 1 \\ \hline
Critic loss coefficient $\omega_v$ & 0.5 \\ \hline
Invalid action loss coefficient $\omega_{invalid}$ & 0.5 \\ \hline
Number of epoch & 10 \\ \hline
Number of processes & 32 \\ \hline
Mini batch size & 512 \\ \hline
Imitation learning timestep & (3e6, 2e6, 0) \\ \hline
Optimizer & AdamOptimizer \\ \hline
Training timestep for the first stage & 7e7 \\ \hline
Training timestep for the second stage & 7e7 \\ \hline
Maximum iterations & (30, 65, 100) \\ \hline

$\mu$ & 20 \\ \hline
$\rho$ &  0.3 \\ \hline
$d_l$ &  1.2 \\ \hline
$d_h$ &  2.2 \\ \hline
\end{tabular}}
\caption{Hyperparameters table}
\label{table:hyperparameter}
\end{table*}

Table~\ref{table:hyperparameter} shows the hyperparameters used to train the MARL model and implement the LNS2+RL algorithm evaluated in Experiment Section of our paper.
The last four hyperparameters are those used to insert the MARL planner into the overall framework, while the remaining hyperparameters are used to train the MARL model. 
There are three different task difficulties in the curriculum learning of training stage one, so \textit{ratio of total number of agents to world size}, \textit{obstacle density}, \textit{$\alpha$}, \textit{Imitation learning timestep} and \textit{maximum iterations} contain three values/value sets, each value/value set corresponding to a task difficulty.
The hyperparameters used in the second stage of training are mostly the same as those in the first stage of training, except \textit{ratio of total number of agents to world size = (60\%, 65\%, 70\%)}, \textit{obstacle density = (15\%, 17.5\%, 20\%)},\textit{$\alpha$ = 0.04},\textit{maximum iterations = 100}

\section{Training Details}
The training was conducted on a server equipped with four Nvidia GeForce RTX 3090 GPUs and one Intel(R) Core(TM) i9-10980XE CPU (18 cores, 36 threads). 
Only one GPU was used during training. 
The code for the neural network part was written in PyTorch 2.1.1. 
We relied on Ray 3.00 to employ 32 processes to collect data in parallel. 
Both training stages one and two lasted 7e7 timesteps, each requiring 136 hours. 
However, the training curve in stage one converged at 4e7 timesteps, and in stage two at 5e7 timesteps. 
We extended the training duration to achieve a model with optimal performance, which can scale to different numbers of agents and adapt to different map structures without retraining.

\section{Summary of Observation Channels}
As an example at timestep $t$, Table~\ref{table:obs} shows the information that agent $i$ will observe.

\begin{table*}[ht]
\centering
\scalebox{0.8}{
\begin{tabular}{l|l}
\hline
\textbf{No.} & \textbf{Information included} \\ \hline
1 & Positions of hard obstacles. \\ \hline
2 & Positions of other agents. \\ \hline
3 & Actual/projected positions of observable agents’ goals. \\ \hline
4 & Position of agent $i$'s own goal. \\ \hline
5, 6, 7, 8 & 
\begin{tabular}[c]{@{}l@{}} 
There are four channels in total, each representing a move action (up, down, left, right).\\ 
When considering only hard obstacles, the value is 1 if selecting the corresponding\\ 
action at a given position brings agent $i$ closer to its goal; otherwise, it is 0. 
\end{tabular} \\ \hline
9 & \textit{SIPPS map}. All cells traversed by agent $i$'s SIPPS path $\hat{p}_i$ during $t-15$ to $t+15$. \\ \hline
10 & 
\begin{tabular}[c]{@{}l@{}} 
\textit{Occupancy duration map}. The ratio of time a cell will be occupied by obstacles\\ 
to the predefined episode termination length. 
\end{tabular} \\ \hline
11 & 
\begin{tabular}[c]{@{}l@{}} 
\textit{Blank duration map}. The ratio of time a cell will remain obstacle-free\\ 
to the predefined episode termination length. 
\end{tabular} \\ \hline
12, 13, 14, 15, 16, 17, 18, 19, 20 & 
\textit{Future path maps of soft obstacles}. The positions of soft obstacles from $t$ to $t+8$. \\ \hline
21, 22, 23, 24, 25 & 
\textit{Predicted path maps of other agents}. The predicted positions of other agents from $t+1$ to $t+5$. \\ \hline
26 & 
\begin{tabular}[c]{@{}l@{}} 
\textit{Cell utilization map}. The normalized sum of the number of timesteps soft obstacles\\ 
occupy the cell from $t-2$ to $t+15$, and the number of timesteps agents occupy the cell\\ 
from $t-2$ to $t$.
\end{tabular} \\ \hline
27, 28, 29, 30, 31 & 
\begin{tabular}[c]{@{}l@{}} 
\textit{Direction utilization maps}. There are five channels in total, each representing an action\\ 
(up, down, left, right, stay). The normalized sum of the number of timesteps soft obstacles\\ 
traverse the cell and select the corresponding action from timestep $t-2$ to $t+15$, and\\ 
the number of timesteps agents traverse the cell and select the corresponding action\\ 
from timestep $t-2$ to $t$.
\end{tabular} \\ \hline
\end{tabular}}
\caption{The observation channels of agent $i$ at time step $t$ and their corresponding information.}
\label{table:obs}
\end{table*}

\section{Map Visualization}
\begin{figure*}[h]
\captionsetup[subfigure]{aboveskip=-1pt}
\centering
\begin{subfigure}[b]{0.17\textwidth}
    \includegraphics[width=\textwidth]{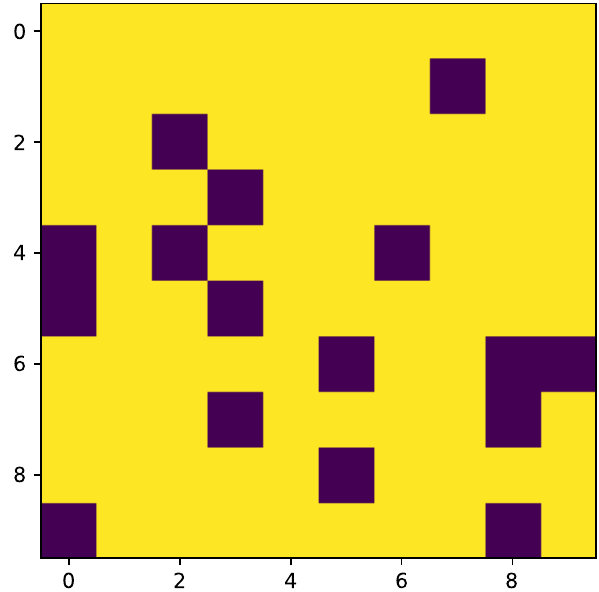}
    \caption{Random-small}
\end{subfigure}
\begin{subfigure}[b]{0.17\textwidth}
    \includegraphics[width=\textwidth]{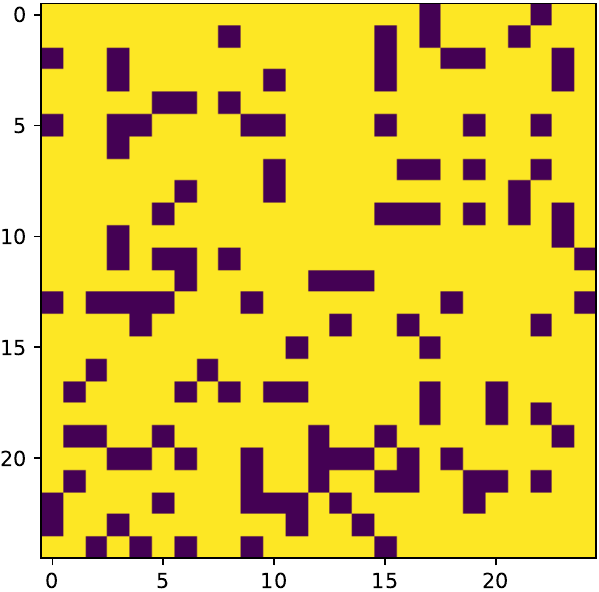}
    \caption{Random-medium}
\end{subfigure}
\begin{subfigure}[b]{0.17\textwidth}
    \includegraphics[width=\textwidth]{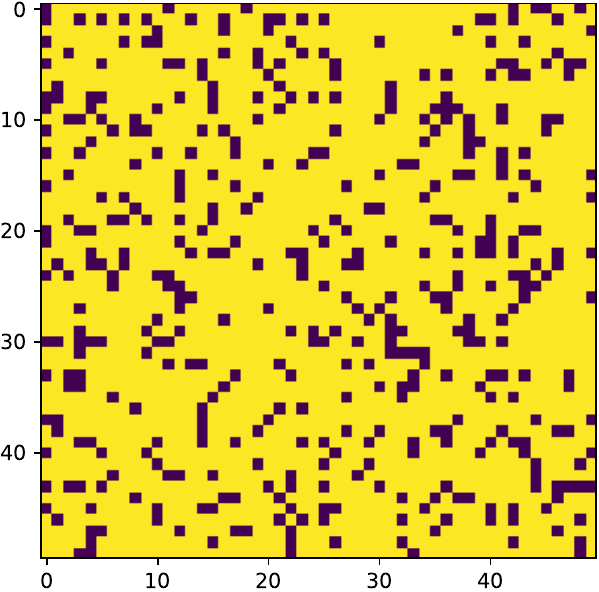}
    \caption{Random-large}
\end{subfigure}
\begin{subfigure}[b]{0.17\textwidth}
    \includegraphics[width=\textwidth]{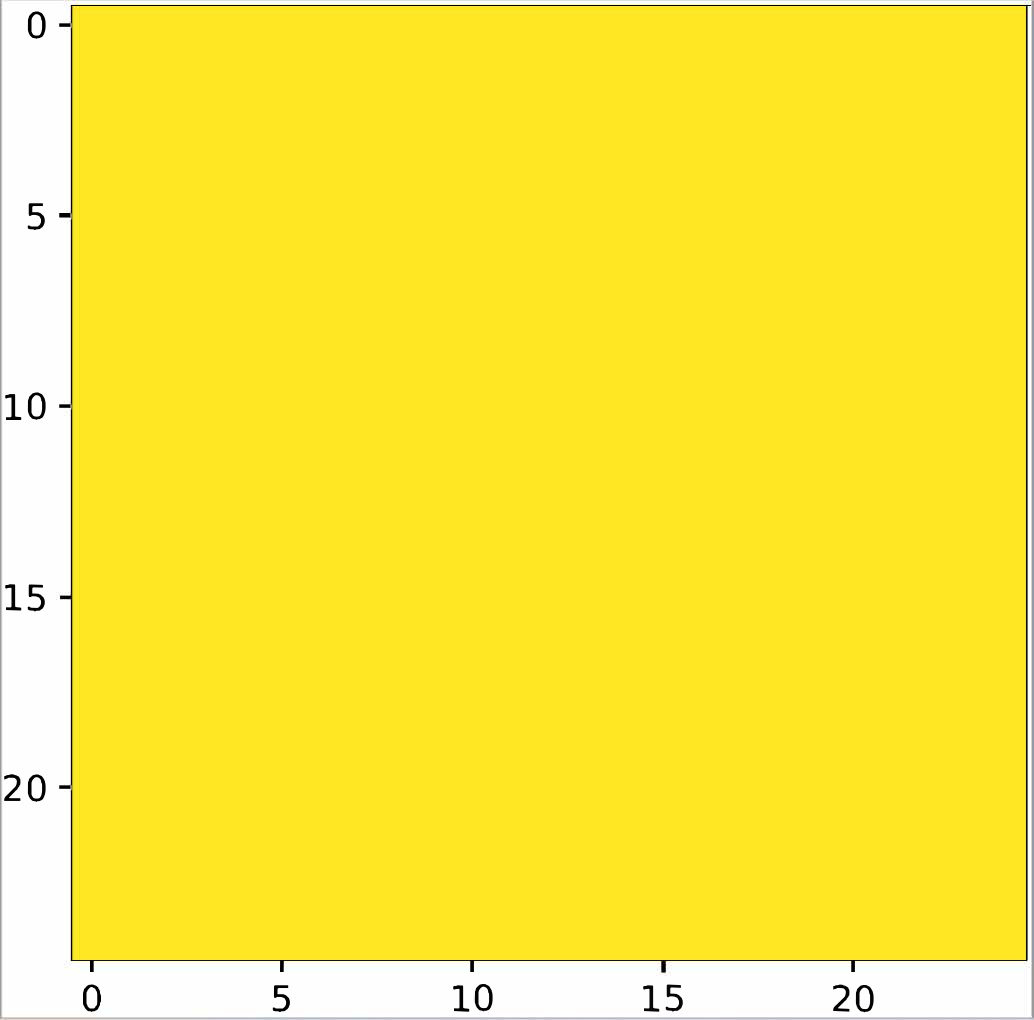}
    \caption{Empty}
\end{subfigure}
\begin{subfigure}[b]{0.17\textwidth}
    \includegraphics[width=\textwidth]{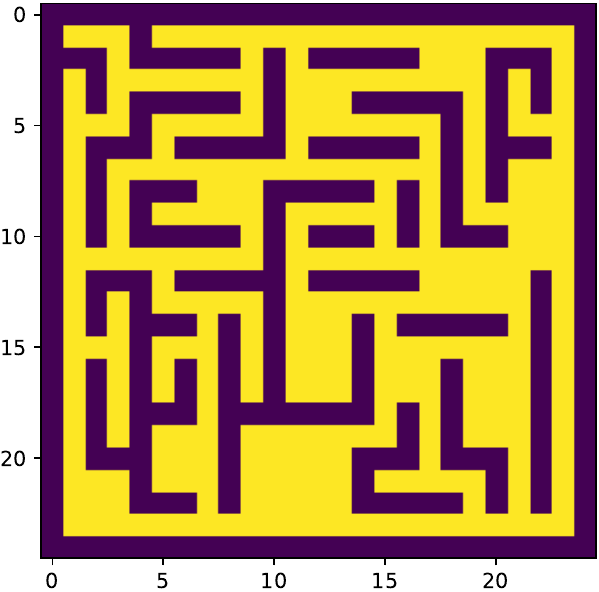}
    \caption{Maze}
\end{subfigure}
\begin{subfigure}[b]{0.17\textwidth}
    \includegraphics[width=\textwidth]{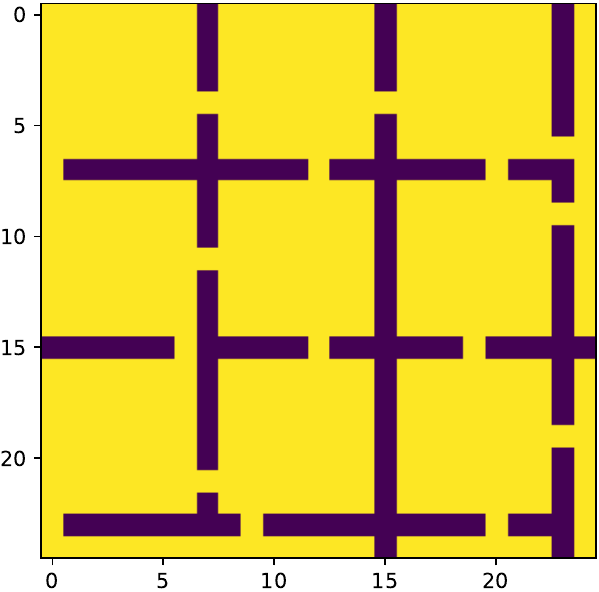}
    \caption{Room}
\end{subfigure} 
\begin{subfigure}[b]{0.17\textwidth}
    \includegraphics[width=\textwidth]{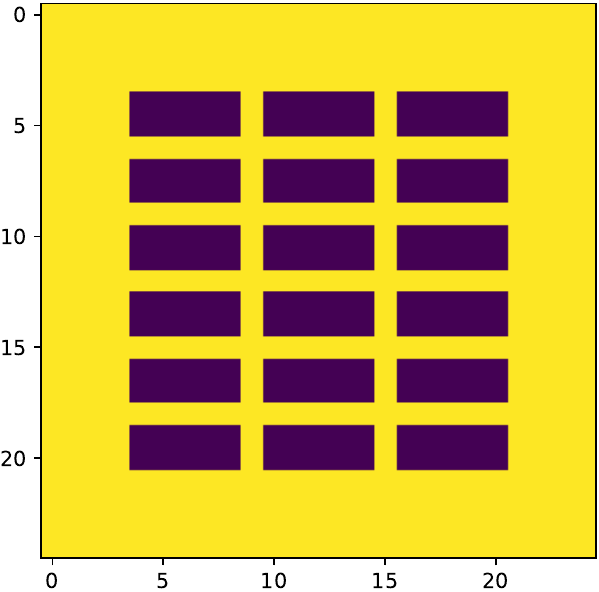}
    \caption{Warehouse}
\end{subfigure}
\setlength{\belowcaptionskip}{-0.5cm}
\caption{Visualized examples of the maps used for empirical evaluation.}
\label{fig:vis}
\end{figure*}

Figure~\ref{fig:vis} illustrates examples of the maps used for testing. 
In all maps, purple cells represent static obstacles, and yellow cells represent empty areas.
The static obstacle layouts of the maze, room, and warehouse maps are fixed, and the obstacle layouts of the three random maps are randomly regenerated in each instance.
For every instance of all maps, the agents' distinct starting and goal positions are randomly chosen from empty cells within the same connected region of the map.
The tasks tested in Experiment Section are challenging, with the vacancy rates for the most difficult tasks on the seven maps only being 17.5\%, 17.5\%, 17.5\%, 20\%, 26.74\%, 25.48\%, and 21.2\% respectively.

\section{Real World Experiment}\label{app_real}
\begin{figure}[h]
    \centering
    \includegraphics[width=0.7\linewidth]{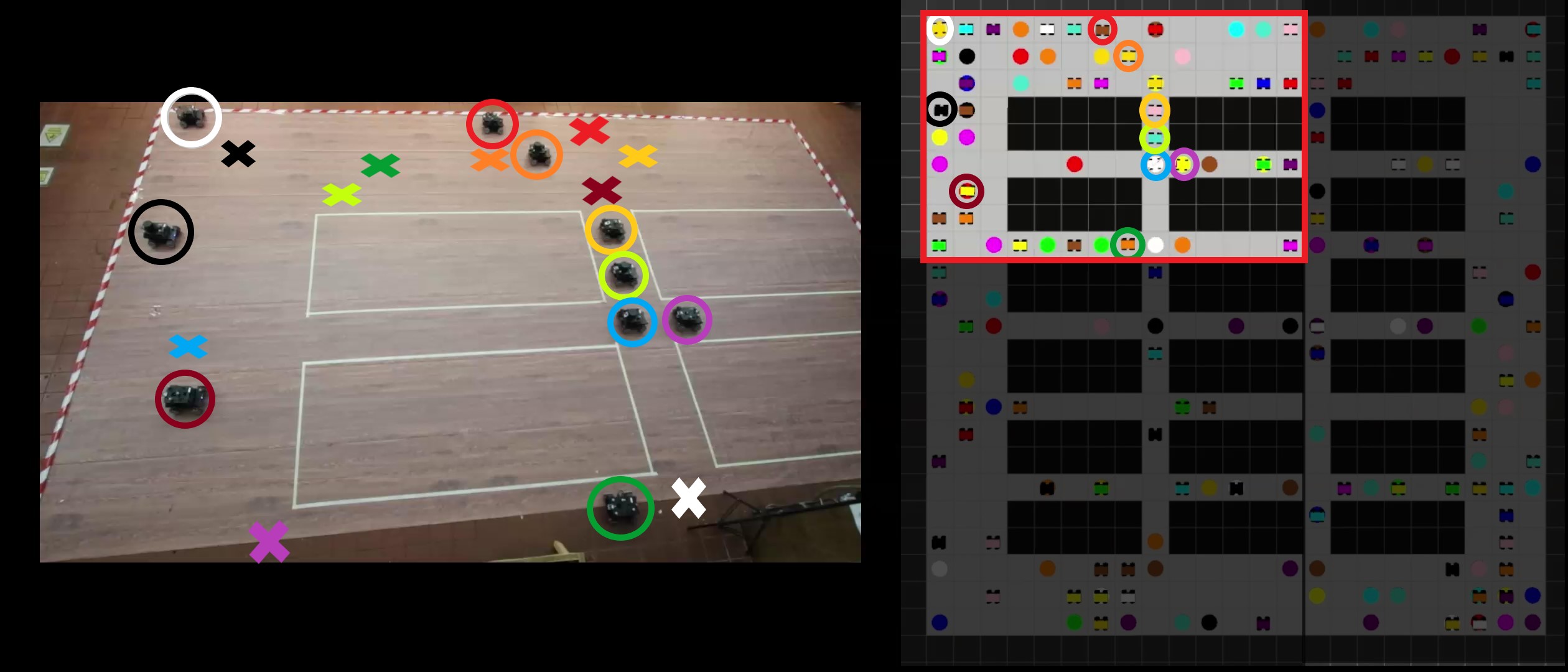}
    \caption{Real portion of the warehouse with 10 robots (left), and complete rviz visualization with 100 robots (right). 
    Real robots are circled in different colors, and crosses of the same colors represent their goal positions
    }
    \label{fig:exp_start}
\end{figure}

\subsection{Setup}
We implemented our algorithm on a team of 90 virtual and 10 real robots in a warehouse environment  (Figure \ref{fig:exp_start}).
We assume that each pixel has a resolution of 0.5 m per grid unit, and based on this resolution, we realize a \textit{7m x 4.5m} part of the warehouse in the real world. 
The 10 real robots are \textit{Sparklife Omnibots} robots with dimensions of approximately \textit{0.3m ×0.3m} equipped with mecanum wheels for omnidirectional movement.
We randomly generate the start and end positions of robots in the environment, with the constraint that the start and end positions of the real robots are in the real part of the warehouse 
Since the MAPF problem assumes the accurate position of all agents for all time, we considered the ground truth positions for all virtual robots and used the \textit{OptiTrack Motion Capture System} to get accurate positions for our real robots. 
Finally, we pre-plan the complete path set using LNS2+RL before actual execution and compute the joint actions agents should take at each timestep based on this path set.

\subsection{Action Dependency Graph}
The paths generated by LNS2+RL assume perfect operation for each agent at every step in the discrete map.
However, due to the robots being imperfect and the environment being continuous, executing the planned path directly in the real world is infeasible.
For instance, a planner may output a set of actions that lets an agent (say 'A') take another agent's (say 'B) position while Agent B moves to a new cell. 
If we were to try and execute this behavior on robots directly, a number of things, like errors in localization, delays in executing motor control commands, or differences in velocities, could lead to a collision between agents.
To avoid such scenarios and make our joint set of actions executable, we follow the method proposed by \cite{honig2019persistent} and construct an \textit{Action Dependency Graph} (ADG).
The main benefit of having an ADG is that it introduces a precedence order for the agents occupying a cell. 
This means that if an agent moves faster than the remaining agents, it will still wait for agents to catch up if it occupies a cell after another agent in the originally planned path. 
This behavior ensures that the errors in executing the planned path do not propagate and hamper the execution plan.
Returning to our earlier example, using an ADG ensures that since Agent A is headed to a cell that Agent B occupies before it is in the originally planned path, it will wait for Agent B to reach a new cell before moving to occupy it.

\subsection{Execution}

To implement the ADG, we first need to convert each action into a task that is to be executed by a robot.
For this, we define a \textit{Task} object as follows:
\begin{Verbatim}[fontsize=\scriptsize]
object Task {
taskID;  // Unique identifier for the task
robotID; // Identifier for the robot which does this task
action;  // Action to be performed by the robot
startPos;//start position of the robot before action execution
endPos;  // Position of the robot after action execution
time;    // Timestep when action to be executed
status;  // Marks if the task is staged, enqueued, or done
};
\end{Verbatim}

During execution, we first cycle through all agents, convert their actions into task objects, and use these tasks to construct an ADG. 
There are three statuses that an agent can have, namely:\textit{STAGED}, \textit{ENQUEUED} and \textit{DONE}.
All tasks are initialized with a \textit{STAGED} status. 
The status is changed to \textit{ENQUEUED} after all their dependent tasks in the ADG are done, meaning that this task is ready for execution. Lastly, the status is marked \textit{DONE} when the agent reaches the \textit{endPos} of the task.
\begin{figure}
    \centering
    \includegraphics[width=0.5\textwidth]{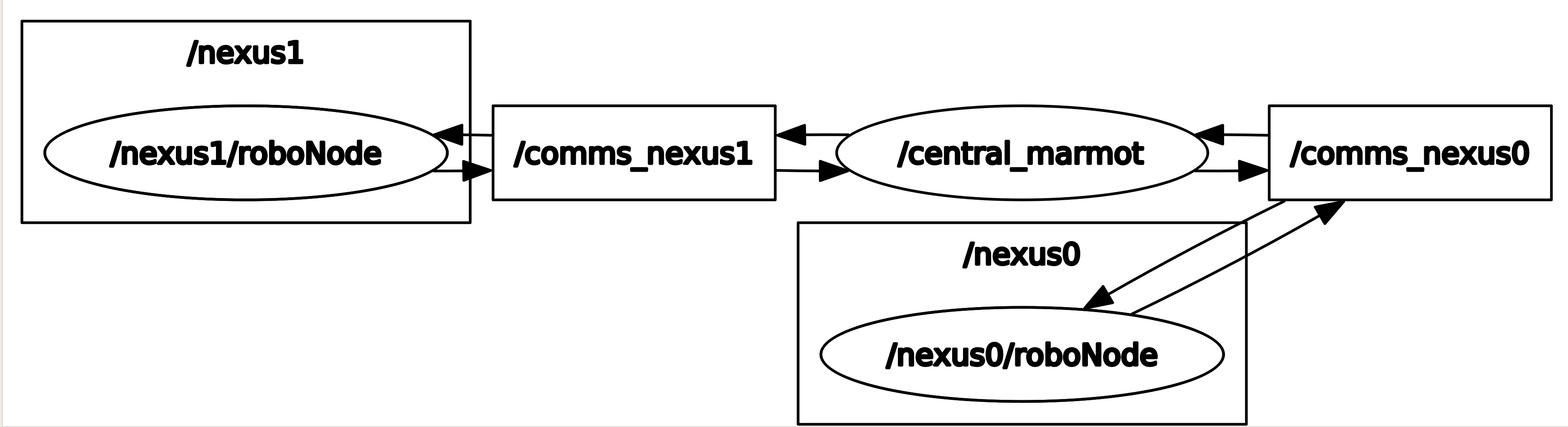}
    \caption{An RQT\_Graph Snippet of our ROS architecture for 2 robots. Ovals indicate nodes, rectangles without ovals indicate topics, and rectangles with ovals indicate namespacing for multi-robots. The central\_marmot node assigns tasks to robots (here nexus0 and nexus1) based on an \textit{Action Dependency Graph}. Robots use those tasks to enqueue waypoints to the next goals. Upon reaching a goal, a robot will relay this information back to the central node.}
    \label{fig:exp_ROS}
\end{figure}

\subsection{ROS Architecture}
We have a \textit{central node} that is responsible for generating and maintaining the ADG and distributing tasks for robots when they are to be enqueued.
Furthermore, each robot has a \textit{robot node}, which is responsible for receiving a task or set of tasks from the central node, extracting a goal from it, and using a PID controller to reach the goal.
When a robot reaches its goal, it sends this information to the central node, marks this task as \textit{DONE}, and uses the ADG to enqueue additional tasks.
Figure \ref{fig:exp_ROS} portrays a snippet of the architecture for 2 robots.

\bibliography{aaai25}

\end{document}